# A Chain-Detection Algorithm for Two-Dimensional Grids

Paul Bonham[1] and Azlan Iqbal[2]

ABSTRACT

We describe a general method of detecting valid chains or links of pieces on a two-dimensional grid. Specifically, using the example of the chess variant known as Switch-Side Chain-Chess (SSCC). Presently, no foolproof method of detecting such chains in any given chess position is known and existing graph theory, to our knowledge, is unable to fully address this problem either. We therefore propose a solution implemented and tested using the C++ programming language. We have been unable to find an incorrect result and therefore offer it as the most viable solution thus far to the chain-detection problem in this chess variant. The algorithm is also scalable, in principle, to areas beyond two-dimensional grids such as 3D analysis and molecular chemistry.

## 1 INTRODUCTION

The algorithm described in this article was designed for simplicity of input requirements. Thus, the only inputs to the initial C++ function that begins the process of valid chain detection are two C++ strings. The first is a specially-constructed text string known as a Forsythe-Edwards Notation (FEN) string describing the chess position (most chess software programs can generate this string from any legal chess position). The second is a string containing the board square (in coordinate notation) of the piece that moved last in the position. For example, if a queen on the d6 square moved last (from another square), the second string would be "d6". The reason for this is that any valid chain as used in SSCC must include the piece that just moved. A 'chain' in Switch-Side Chain-Chess (SSCC) is basically a connected sequence of pieces (color is irrelevant) surrounding at least two empty squares. Figure 1 shows four examples of valid chain formations. A complete description of what constitutes a chain in SSCC and more diagrams illustrating them can be found in (Iqbal, 2013, 2014a).

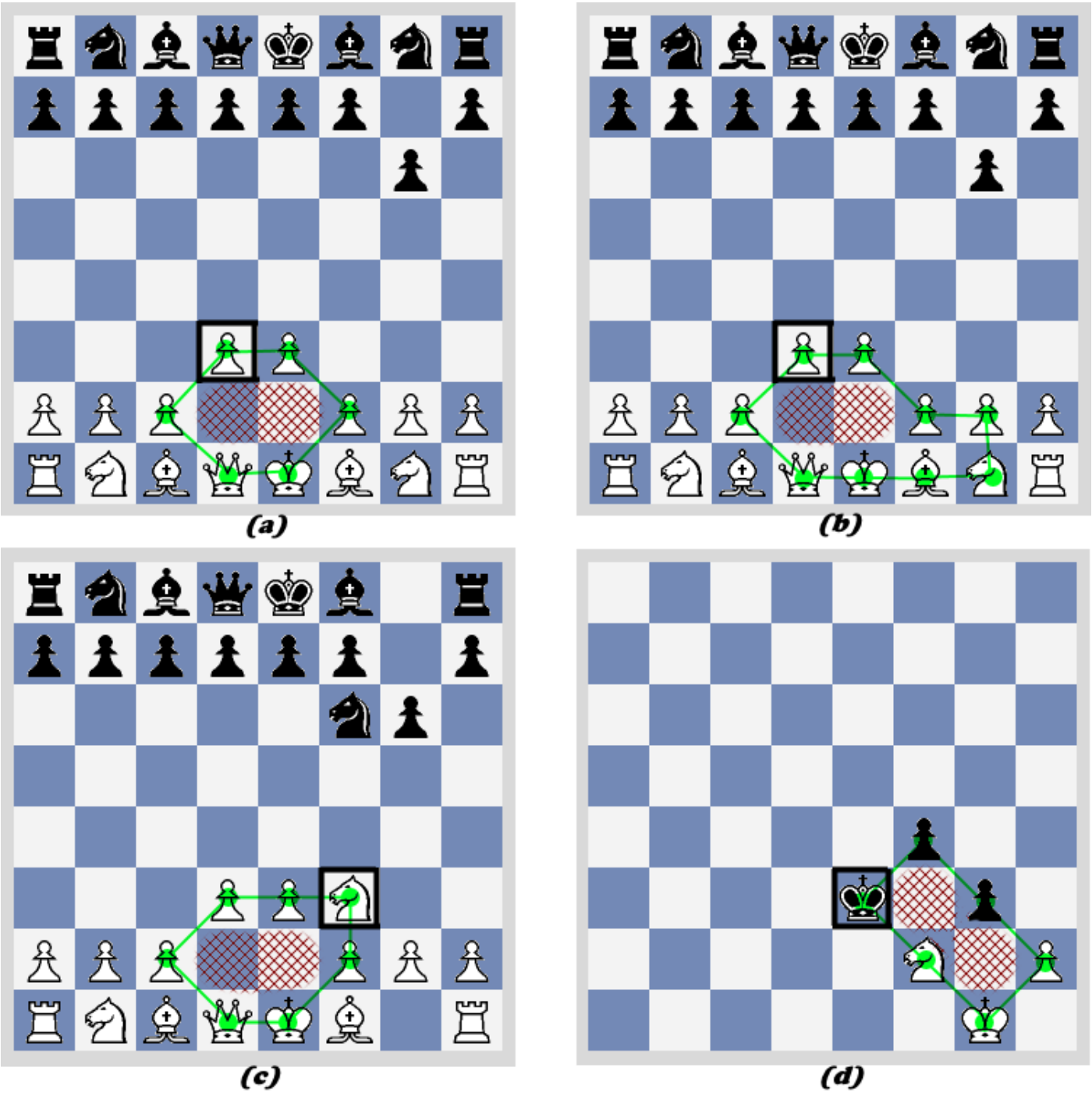

**Figure 1:** Four examples of valid chain formations in SSCC.

[1] *Self-employed, freelance C++ contractor; pwb@twitoftheyear.com*

[2] *College of Computer Science and Information Technology, Universiti Tenaga Nasional, Putrajaya Campus, Jalan IKRAM-UNITEN, 43000 Kajang, Selangor, Malaysia; e-mail: azlan@uniten.edu.my*

The piece that just moved to form the chain is shown with the square highlighted with a black border. In (a), we see the 2-empty-square minimum has just been met. In (b) is shown an alternate chain formation that can be perceived in the exact same position and involving the same 2 empty squares. There are many more that form the required Euler path. In (c), White has made another move to form a brand-new chain that didn't exist in (a) or (b), enclosing the same 2 empty squares. In (d) we see the minimum number of pieces (6) needed to form a chain around 2 empty squares, which in this case are connected diagonally. Section 2 presents our methodology in detail. Section 3 explains a sample case of chain detection using a particular chess position. We briefly discuss some outstanding issues in section 4 and conclude our article in section 5 with some directions for further work.

## 2 METHODOLOGY

The algorithm begins by parsing the FEN string to populate an array of 64 C++ strings, one for each square on the chess board. Each string thus represents whatever piece happens to occupy that square, an example being if it is a white knight on the square, the string would be "WN" (N is used for Knight because K is for the King). If the square is unoccupied, the corresponding string is left empty. The algorithm does not actually need to know which piece is on a square, only whether it is occupied or not. That Boolean value is captured by the content-state of the string (i.e. empty or not empty). Other pieces of information about each square that *could* have been captured within a C++ data structure such as a class were deemed unnecessary for chain detection. For example, whether a certain chessboard square is 'under attack' is meaningless in chain detection. All that matters is the connected structure on the two-dimensional grid. This makes the algorithm potentially applicable to a wide variety of board games and also other areas where grids are used.

Any error in the parsing of the input FEN string will produce an output message that the string is invalid. The algorithm is currently limited to an 8 x 8 chess board, but can be easily be modified to include other dimensions of square or rectangular boards. The width and height in squares could be made into two further inputs to eliminate the need to change the code significantly and recompile the software. If the board dimensions are changed then the input FEN string would have to change accordingly as well. Although the array of C++ strings is 64 elements in size, it is processed as an 8 x 8 two-dimensional array representing the board. Also parsed is the second input string in order to convert the board square notation (such as "d6") into an index number in the 8 x 8 array, i.e. an index between '1' and '64'. This number will be stored for later use by the algorithm.

In previous work (Iqbal and Salih, 2012), the method described takes the position of the last-moved piece as a starting point or root. A search for pieces is then done of that location's (usually 8) immediately surrounding squares, storing those locations in a 'working queue' and then visiting elements in the working queue and successively making them new roots in a recursive process which also involves moving visited elements from the working queue into a 'visited queue'. Each recursive visit increments the working 'level' by 1. The method, however, admittedly causes the occasional false positive or false negative. In this work, we focus instead on what could be considered a reverse technique where the idea is not on what a chain *is* but rather what it *is not*.

The first pass through each element of the array performs '8-connected component analysis' (8-CCA)[3] – not of the starting point and outward looking for connected occupied squares, but of the entire board looking for connected *empty* squares. This enumerates and classifies any and all separate groups of connected empty squares, each set of which shall henceforth be referred to as an 'EA' or 'Enclosed Area'. Note that 8-CCA is a known technique and is not presented here as being new (Fisher, Perkins and Walker, 2003). The classification of separate islands of connected empty squares is simple integer counting, i.e. if there are 3 separate EAs in the chess position, the first will have all its enclosed squares labeled as "1", the second as "2", and the third as "3". A 64-element integer array stores this 'class' value for each board square that is part of an EA. Figure 2 shows an example of this integer classification of EA's in a given chess position.

In the chess position below, the red cross hatching shows the three separate valid EAs. Each is given its own integer class value as part of the 8-CCA process after a filtering routine has removed all other connected empty squares as being invalid for an SSCC EA. In the second pass through this integer array a filtering process is performed in which any elements that are, a) on the rim of the chessboard or (b) vertically or horizontally connected to empty squares on the rim of the chessboard, get their integer value set back to 0.

---

[3] *Connected Component Analysis is a technique used in computer vision to detect connected regions in binary digital images. A reasonably good explanation of it can be obtained freely at https://en.wikipedia.org/wiki/Connected-component_labeling*

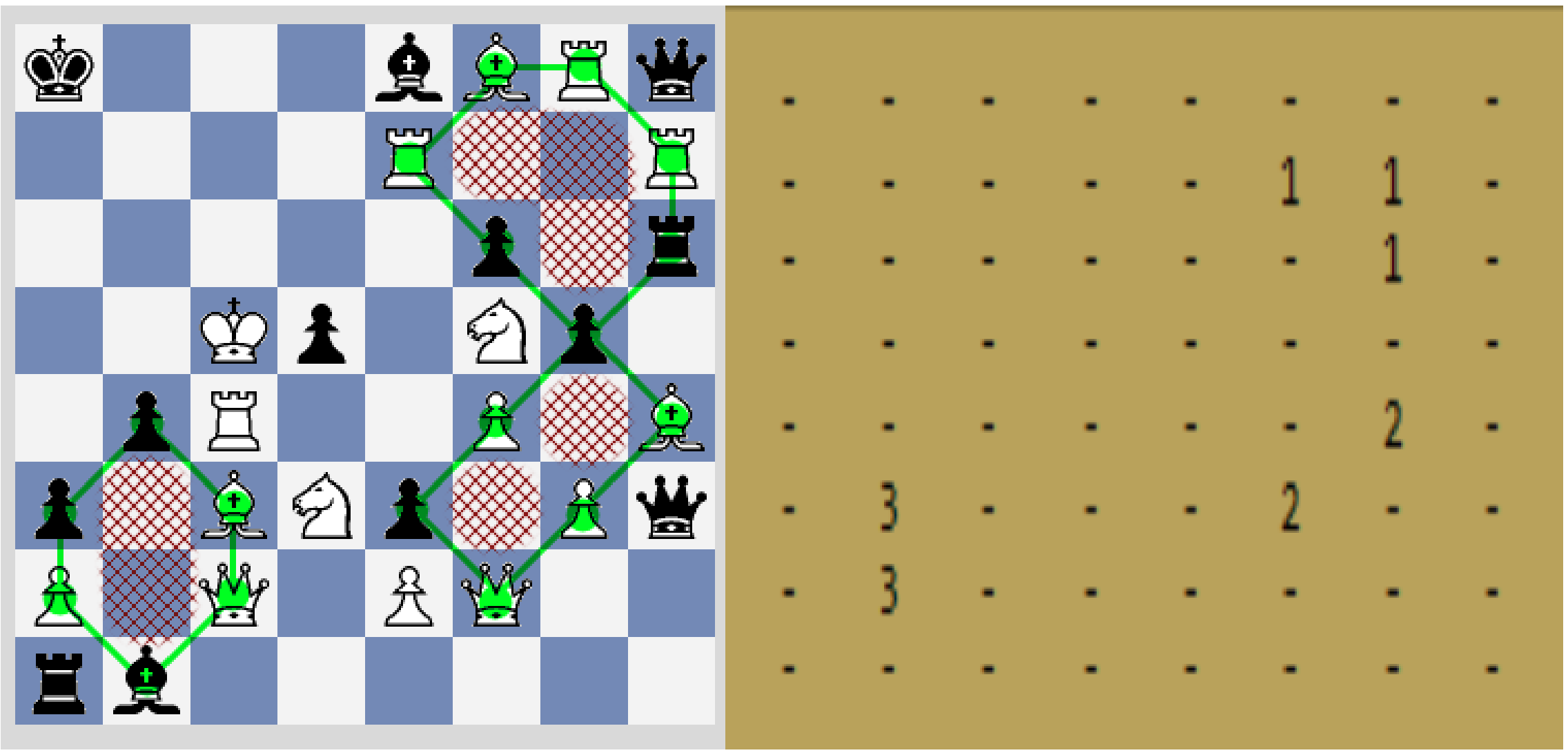

**Figure 2:** An example of an 'EA Class Count'.

This effectively eliminates those nodes as being part of any EA. The reason is because a valid SSCC chain, as described in the rules, cannot enclose an island of empty squares that reach to the rim of the chessboard by horizontal or vertical connection (see Figure 3(a)). This is what is meant by the algorithm using the idea of what a chain 'is not'. However, an island of empty squares that only connects *diagonally* to one or more empty squares on the rim of the chess board (Figure 3(b)) may be enclosed by a valid SSCC chain so such diagonal connections are left as they are. Regardless, all empty squares on the rim itself are eliminated irrespective of their connections to other empty squares.

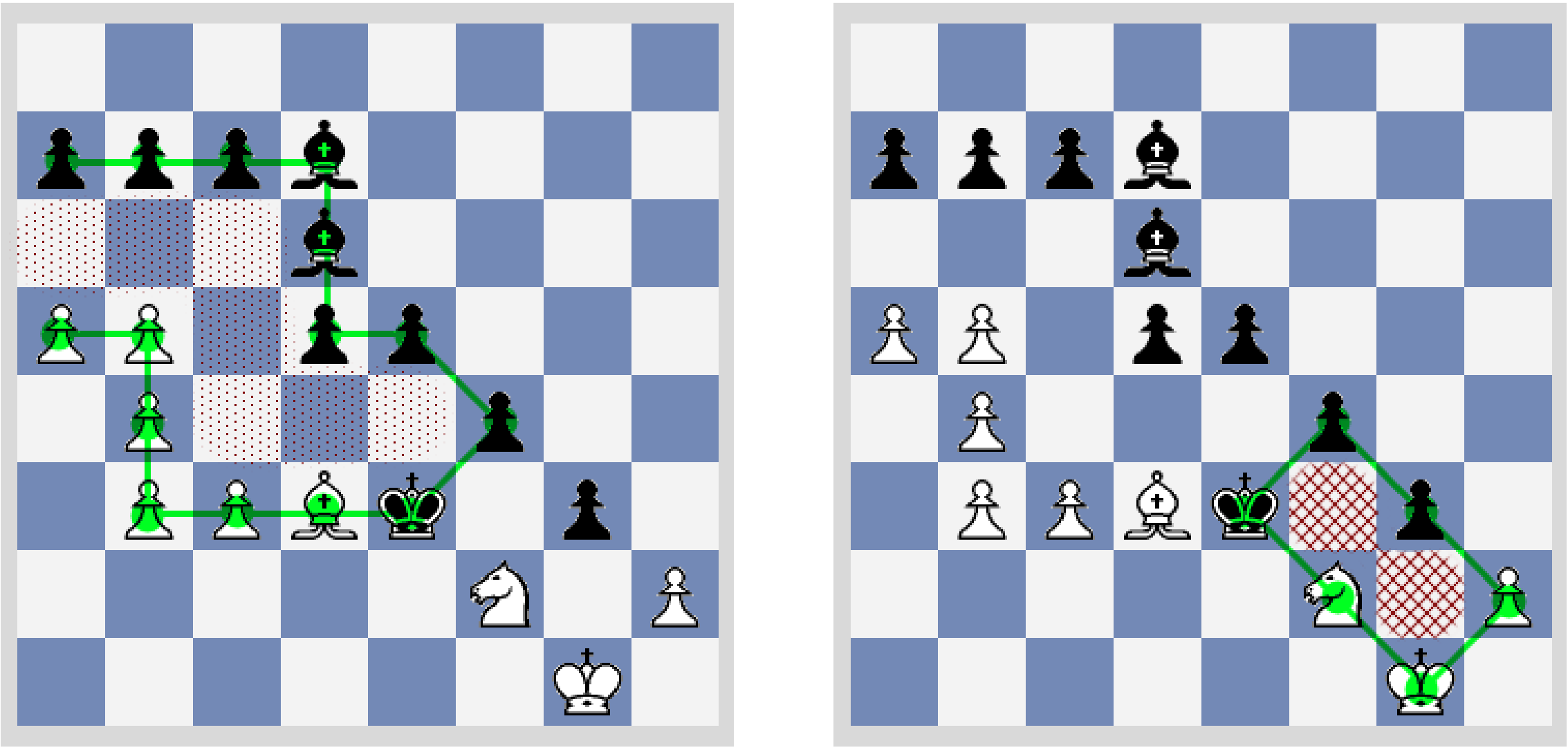

(a) (b)

**Figure 3:** Square on the rim.

In Figure 3(a) we see that the path of dotted empty squares leads horizontally and vertically to the rim of the chessboard and the chain of pieces cannot be closed. The filtering routine thus will eliminate these squares from being part of an EA. However, in (b) we see that a diagonal path to an empty square on the rim does NOT eliminate the 2 empty squares shown with cross hatching from being part of a valid EA because a Euler path of connected pieces can enclose them. A third pass through the integer array eliminates any EAs that only have a single node or empty square (because a valid chain in SSCC must enclose at least 2 empty board squares). The function that does this elimination takes the minimum size required for a chain as a parameter, so that if SSCC were to change its rules or have sub-variants of the game introduced (requiring, say, a valid chain to have 3 enclosed empty squares), the algorithm could handle this as well with a simple modification.

A fourth and final pass through the 8 x 8 array does any necessary final correction of classification of group numbers. For example, if the first two passes identified five separate EAs, and then the third pass eliminated the 4th EA (classified as '4') because it had only a single empty square, the EAs left would be classified '1', '2', '3', '5'. The final pass would therefore correct the final EA ('5') and re-classify it as '4'. Technically this is not necessary as the classification is somewhat arbitrary and serves only to differentiate EAs from each other, but this correction is nevertheless good practice. When the final phase of the algorithm takes effect, it will be searching for two unique paths from the last-moved-piece's board square to *the same* EA. Note that two unique paths going to two *different* EAs do not validate a chain.

## 3 A SAMPLE CASE

Figure 4 shows a diagram of a chess position that will be used to create a representation of what the process as described so far will produce. Please ignore that this chess position is, in fact, an impossible albeit legal one.[4]

The FEN string for this position is:
*1n1k1b2/rpp1pp2/2b1pn2/rq2pp2/1Q4PB/PPPNP1PB/3P1R1P/3K1RN1 w - - 0 1*

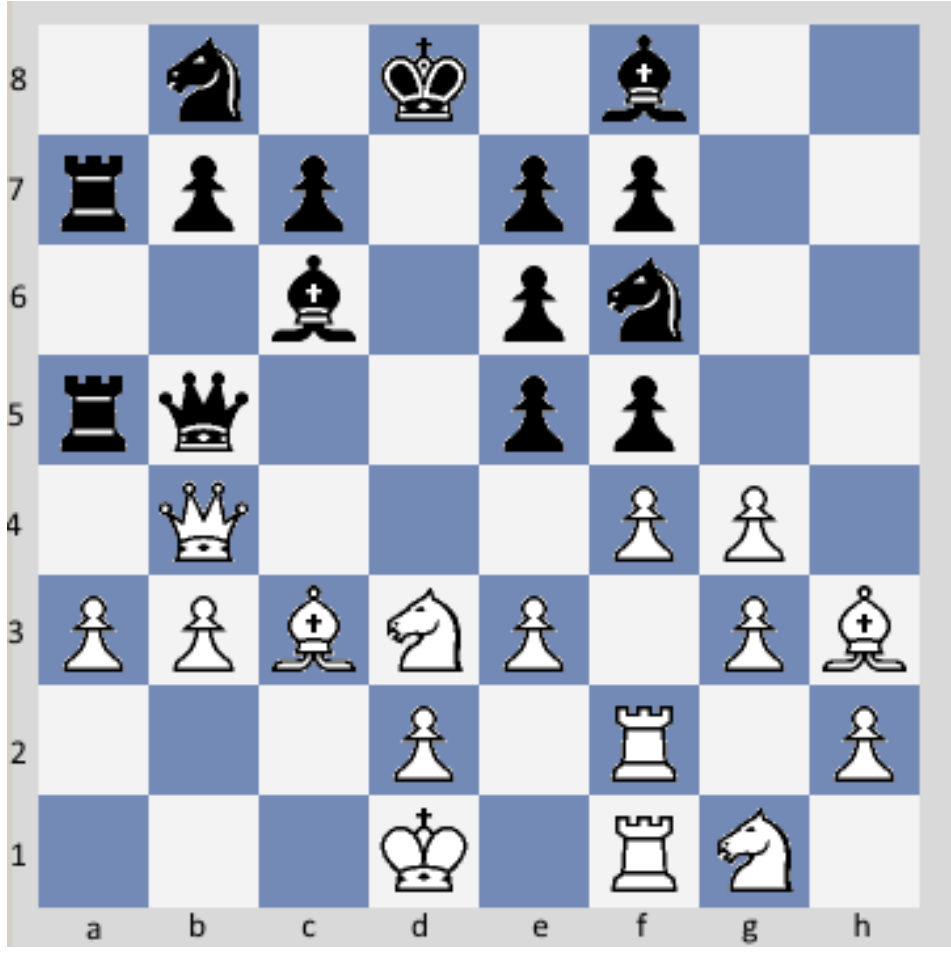

**Figure 4:** A sample position

Figure 5 demonstrates what the first, second and third (if applicable) passes through the array produce. The squares with no marking are empty but *ineligible* to be part of any enclosed area (EA). In this case, it is because they are either on the rim of the board or vertically or horizontally connected to such squares. The squares marked with a solid black circle are occupied by pieces. Only the squares marked by the crisscrossing lines are both empty and eligible to be part of an EA. There are a total of only 9 such squares in this position, and in this case they are all connected such that this is all considered one group of connected squares. Looking back at the actual chess position shown in Figure 4, it may not be immediately obvious to the human eye, hence the novelty of chain 'detection' in the SSCC variant. Thus all of the valid empty squares are classified as group 1 by the 8-Connected Component Analysis (8-CCA) step mentioned in the previous section. However, in reality, there are actually four separate possible EAs in this particular example as Figure 6 shows.

---

[4] *There are countless positions in chess that are "legal", meaning they do not violate rules of the game but are nevertheless "impossibie" because no sequence of legal moves could lead to the position. In this example, the positions of the Black pawns could only happen if the pawns captured opposing pieces but all 16 of White's pieces are on the board. White's pawn on g4 is also an impossibility. Discovering such impossibilities would be a part of what is known in chess as 'retrograde analysis' (Smullyan, 2012).*

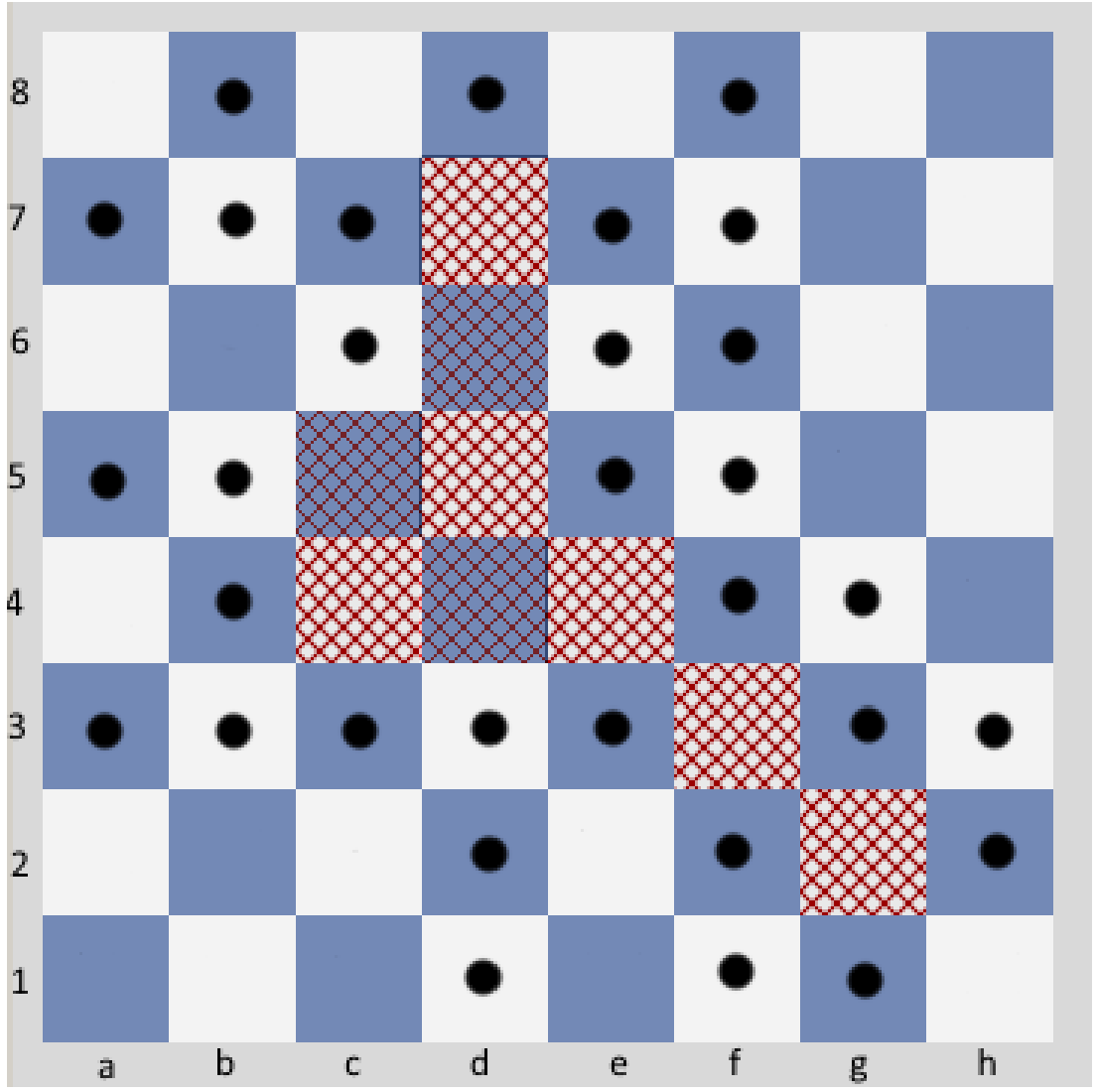

**Figure 5:** The valid EA squares remaining (marked with cross hatching) after three filtering passes.

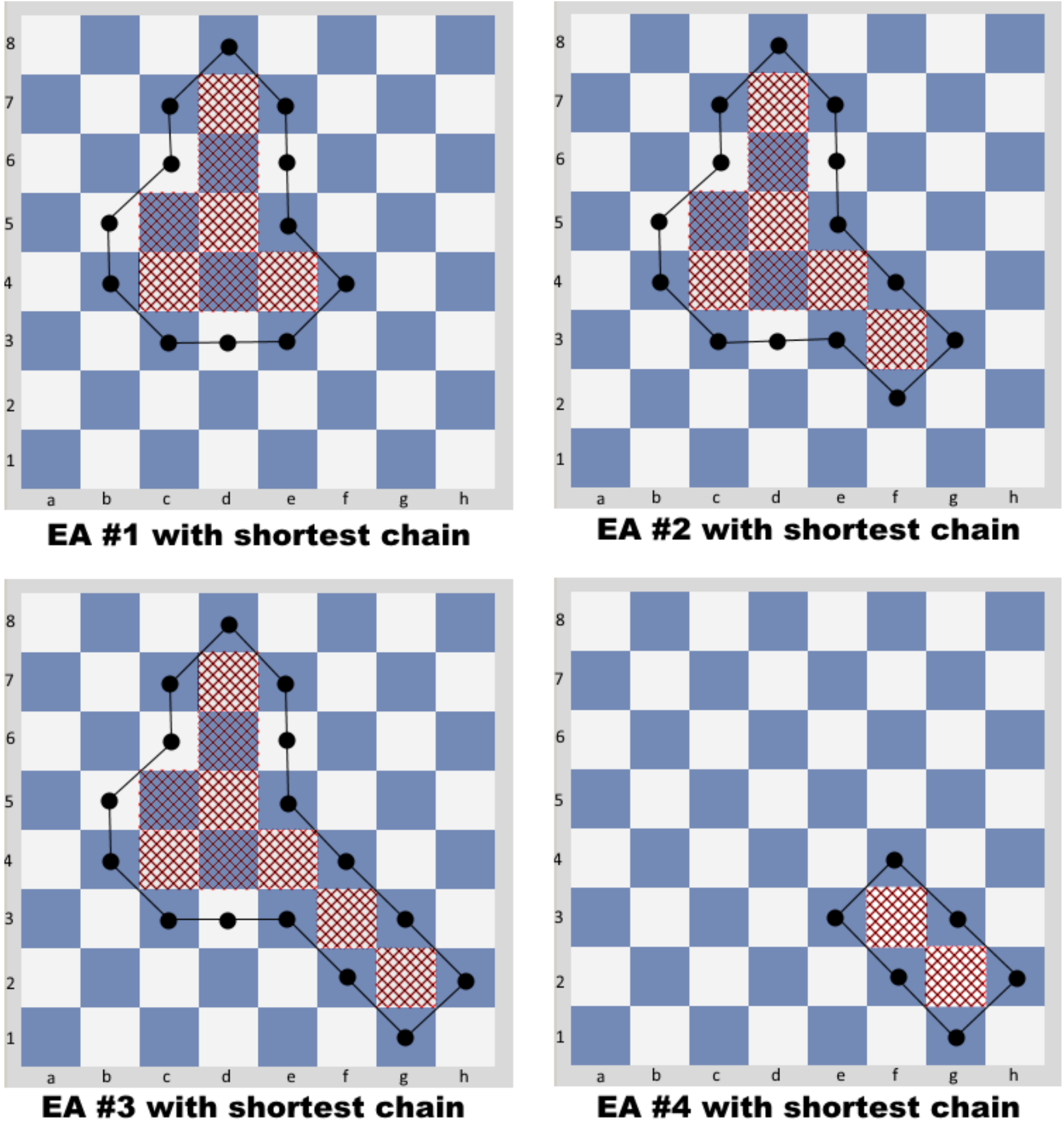

**Figure 6:** Four possible valid SSCC chains from the position in Figure 4.

For each of the four possible EAs, we show the shortest or smallest possible chain that can surround it. However, these shortest chains shown as black dots connected by lines are by no means the *only* chains that can surround each EA. The smallest possible EA, in SSCC, would be as shown in Figure 6 (EA #4). This is because of chains as they are defined in SSCC, which states that a chain must surround a minimum of two connected empty squares. It therefore takes a minimum of 6 pieces to form a chain, so the smallest possible chain in this case would be comprised of the pieces on e3, f4, g3, h2, g1 and f2 (the black dots in EA #4). In this position, we also simply cannot add the e4 square to the f3 and g2 enclosed ones and declare that these three squares make up another separate EA. This is because in order to enclose e4, f3 and g2 with a chain *given the pieces actually on the board* we have to add more empty squares until we have added all the rest of them, as shown in Figure 6 (EA #3).

This is also the case if we begin with the EA squares d7 and d6. They are two connected EA squares but there is no chain of piece-occupied squares surrounding just d7 and d6. So again we must add more EA squares to get a chain, until we have added all except f3 and g2, so we can form the chains as shown in EA #1. By the way, EA #2 is just EA #1 with the f3 square added in the enclosure. It is valid because a chain can be formed around it as shown by the connected black dots, specifically the f2 and g3 ones.

In any case, the algorithm for chain detection is not concerned (at least not for SSCC engine purposes) with actually finding all possible chains for a given EA. It only needs to answer the question in a Boolean fashion (true or false), i.e. is there a *valid* chain around any one of the EAs found by the 8-CCA?

In order to answer that, one first needs to confirm the location of the last-moved piece, since SSCC rules state that only a chain that includes it qualifies (a clever rule by the way, as it insures that such a chain, if it exists, is newly-formed and not pre-existing). The rules of SSCC with respect to valid chains also state that each node of the chain can only have one input connection from another node and one output connection to another different node (thus ensuring a Euler path). This basically means that for any piece-containing board square (that is not directly adjacent to an EA square) to be part of a valid chain around a given EA, there must be two or more unique paths via piece-containing squares that never use the same such square twice to get to any EA square that is included in the given EA. For example, in Figure 5, consider the square a3. It has only two possible paths of squares with pieces on them (black dot) to follow to reach an EA square, i.e. b4 and b3. Both lead directly to a valid EA square node that is the c4 square. Therefore a3 has two unique paths to one and the same EA, and thus can be part of a chain.

This is where the algorithm gets into a rather tricky and somewhat complex area. It must determine whether the square occupied by the last piece that moved has at least two unique paths to the same EA block of squares. The example just given was very easy to spot but sometimes even a human can have trouble determining this when there are multiple pieces in varying arrangements separating the last-moved-to square from the nearest empty square that is part of an EA. It this can be difficult for humans, a computer algorithm need to be sufficiently complex and robust to be able to work through the possibilities.

There are 4 main stages to the overall algorithm.

- input validation and parsing
- 8-CCA to classify EAs (if any)
- filtering (which in this case is based on SSCC rules, but could also be anything else)
- attempt to find two unique paths from last-moved-to square to a single EA

The process will stop, declaring no valid chains found, if it determines there is invalid input, or if there are no EAs in the position, or if after filtering there are no EAs remaining from the ones that were found initially. In this article, we will not describe the parsing of the FEN string which is a fairly straightforward exercise. There are many FEN parsers in existence and they are not difficult to create once FEN notation is understood. A detailed description of the algorithm and pseudocode are provided in Appendix A.

## 4 DISCUSSION

Testing has thus far not identified a position where the algorithm returns a false positive or false negative. However, during examination of 240 test cases earlier on, there were three new ‘corner’ cases discovered with each requiring a change to the algorithm. One of these was relatively major, involving the timing of the search through the candidate paths. More information on this is provided in Appendix B. The main point here is that other corner cases may, at present, be unknown. This is therefore perhaps a classic case of being unable to prove a negative. Only further testing over time by incorporating this algorithm into an actual SSCC engine and perhaps other areas will help to either identify limitations of the algorithm or increase confidence in its effectiveness.

Unfortunately, comparisons with the false positive and false negative cases identified in previous work (Iqbal and Salih, 2012) could not be done: these examples were not documented in the article and eventually lost due to unforeseen circumstances. We are in a similar predicament when it comes to comparing the efficiency of the algorithm against the approach in the aforementioned previous work as no efficiency tests were performed there as well. It is quite possible that the present algorithm is less efficient than the original one given that the one presented here iterates through the entire board array at least three times and then iterates outward from the square moved to by the last piece to move. So the best balance between accuracy and efficiency is difficult to determine here. In an actual SSCC engine, millions of positions would likely need to be analyzed every minute along with the standard set of chess heuristics so efficiency is not an unimportant issue in that case.

A notable limitation of the present algorithm, in its early versions, was its use of recursion when iteration would actually be preferable. Recursion has a built-in limitation on the depth to which the algorithm can continue until an overflow of stack resources[5] occurs. That has not been a problem for the 8x8 chessboard but it would likely be for much larger 2D spaces and certainly if the algorithm were modified for 3D spaces.

A change to iteration was made, and the pseudocode we present further on reflects that change. Replacing recursive calls with inline code made the code less modular and less readable (it even includes 'goto' statements which are considered bad form in programming), but removing the limitations of recursion was well worth that cost..

Regardless, the algorithm as it stands, even if efficiency is poor, works in being able to identify valid chains (in SSCC) with no identifiable errors thus far. Incorporated into an SSCC engine, and working in conjunction with a minimax search that considers the heuristics of switching sides, should make a computer quite adept at playing the game. This is unlike the only existing SSCC computer program (Iqbal, 2014b) which is freely available for download. It cannot play well due to the lack of being able to identify chains with high accuracy and even more so because of being unable to decide whether or not it is prudent to switch sides at any given point in the game if a chain is present (yet another aspect of the chess variant that requires further research).

## 5 CONCLUSIONS

In this article, we have introduced a new algorithm that is able to detect chain formations in a two-dimensional grid. Specifically, it has application in the chess variant known as "Switch-Side Chain-Chess" (SSCC). A previous attempt at developing such an algorithm resulted in something that works but occasionally led to false positives or false negatives. The present algorithm also had to be modified in light of such examples but appears to be stable in its current form, even though we cannot 'prove a negative' in claiming that there are no possible chess positions with chain formations where the algorithm would not fail. Regardless, we believe the algorithm is ready for implementation in any SSCC engine where chain-detection would be essential in order to play well. The heuristics necessary for switching-sides prudently is a separate issue beyond the scope of this article.

The present algorithm is also clearly scalable to larger boards and even an additional third dimension, making its applicability not limited to SSCC or even just board games. It would seem possible that this algorithm could be expanded to use three dimensions by changing the 8 x 8 chess board array into, for example, an N x N x N cubic volume and changing the 8-CCA section to a 26-CCA section (*assuming perfect grid arrangements, a 3 x 3 x 3 volume of square-faced cubes has a total of 27 cubes and the central one is connected to all 26 others*). This would cause a significant computational slowdown but is something that could be compensated for by using concurrency techniques. Board squares or simple cubes could therefore be replaced conceptually by 2D or 3D structures representing perhaps molecular or more abstract elements for application in chemistry or physics.

The number of connections between elements would have to be known for the CCA component of the algorithm to work. The number of 'empty squares' (that for SSCC is set at two) can easily be adjusted in the algorithm and the condition of being 'empty' could also be changed, perhaps by using more comprehensive data structures than the simple C++ string used for SSCC and by changing the input requirements. These adjustments would be relatively easy in most cases given that C++ is very suitable for custom abstract data types. The incorporation of

[5] *A stack overflow is an undesirable condition in which a particular computer program tries to use more memory than the call stack has available. In programming, the call stack is a buffer that stores requests that need to be handled. The size of a call stack depends on various factors.*

C++ templates could even make the algorithm adjustable at runtime to accommodate multiple requirements at once. Improvements and enhancements to the algorithm as we have presented it are nevertheless both welcome and expected in the years to come.

## APPENDIX A: ALGORITHM DESCRIPTION AND PSEUDOCODE

There is a global integer array used throughout the process, referred to in the code as **g_iClassesArray**. This array has 64 elements, indexed by integer from element [0] to element [63]. The very first element is for the upper left board square as seen by the player with the White pieces, which is the board square notated as a8. The following table shows how each board square (looking at the board from the vantage point of the player with the White chess pieces) corresponds to each global array index. As shown in Table 1, the board square is the first entry in each cell, and the second entry after the forward slash is the corresponding index into g_iClassesArray:

**Table 1:** The global integer array.

| | | | | | | | |
|---|---|---|---|---|---|---|---|
| a8 / 00 | b8 / 01 | c8 / 02 | d8 / 03 | e8 / 04 | f8 / 05 | g8 / 06 | h8 / 07 |
| a7 / 08 | b7 / 09 | c7 / 10 | d7 / 11 | e7 / 12 | f7 / 13 | g7 / 14 | h7 / 15 |
| a6 / 16 | b6 / 17 | c6 / 18 | d6 / 19 | e6 / 20 | f6 / 21 | g6 / 22 | h6 / 23 |
| a5 / 24 | b5 / 25 | c5 / 26 | d5 / 27 | e5 / 28 | f5 / 29 | g5 / 30 | h5 / 31 |
| a4 / 32 | b4 / 33 | c4 / 34 | d4 / 35 | e4 / 36 | f4 / 37 | g4 / 38 | h4 / 39 |
| a3 / 40 | b3 / 41 | c3 / 42 | d3 / 43 | e3 / 44 | f3 / 45 | g3 / 46 | h3 / 47 |
| a2 / 48 | b2 / 49 | c2 / 50 | d2 / 51 | e2 / 52 | f2 / 53 | g2 / 54 | h2 / 55 |
| a1 / 56 | b1 / 57 | c1 / 58 | d1 /59 | e1 /60 | f1 / 61 | g1 / 62 | h1 / 63 |

The important thing to note from this table is that progressing through the array from 0 to 63 is like starting from the *top left* board square, *moving right* until the end of that row, *then moving down and to the leftmost* board square and again *moving right*. This process is repeated until arriving finally at the *bottom right* board square. The 8-CCA part of the algorithm is also fairly standard in digital image analysis techniques, and the pseudocode that we have implemented can also be found here: *https://en.wikipedia.org/wiki/Connected-component_labeling#Two-pass*.

Initially, the global array has all its elements set to integer value 0. This represents that each square of the chessboard is *not* part of any EA (Enclosed Area). After the FEN string is all parsed (assuming no errors), any board square that does not have a piece occupying it has its corresponding element in the global array set to integer value 1. The reason the value 1 means "not a piece" is that an EA is a cluster of empty squares, not occupied squares. So for the moment, before 8-CCA is done, every blank square is part of an EA labelled as "1".

The **Do8CCA(...) function** is then run to determine how many separate EAs there actually are, where a separate EA means one that does not have any horizontal, vertical or diagonal connection to another EA. This is critical because the coming chain detection process has to know that two unique chains of pieces from a given starting square each connects to the *same* EA. Figure 7 shows how this works with a sample position. The leftmost section shows the labeling of the pieces in the 8 x 8 string array (once again, looking at the array as if it is the chess board from White's viewpoint, where the a8 square is upper left and the h1 square is lower right). Upper case letters indicate White pieces and lower case letters indicate Black pieces where K or k = King, Q or q = Queen, R or r = Rook, B or b = Bishop, N or n = Knight, P or p = Pawn. The center section shows what the 8 x 8 global integer class array contains before CCA, and the rightmost section shows what that same array holds after the CCA (note: dashes are shown here instead of "0" values for better visibility of the pertinent class numbers). The word "classes" is used in lieu of "EAs" to make the output more generic.

```
                                   Classes Before CCA:                Classes After CCA:

-  k  -  r  -  -  -  -             1  -  1  -  1  1  1  1             1  -  2  -  3  3  3  3
b  p  P  Q  n  R  N  R             -  -  -  -  -  -  -  -             -  -  -  -  -  -  -  -
-  -  -  q  R  -  -  b             1  1  1  -  -  1  1  -             4  4  4  -  -  5  5  -
-  -  n  P  -  -  -  p             1  1  -  -  1  1  1  -             4  4  -  -  5  5  5  -
-  P  P  B  p  B  -  p             1  -  -  -  -  -  1  -             4  -  -  -  -  -  5  -
r  P  -  -  -  -  p  B             -  -  1  1  1  1  -  -             -  -  5  5  5  5  -  -
-  p  Q  -  -  N  -  n             1  -  -  1  1  -  1  -             5  -  -  5  5  -  5  -
K  -  -  -  -  r  -  -             -  1  1  1  1  -  1  1             -  5  5  5  5  -  5  5
```

**Figure 7:** Contents of the global pieces array and global classes array before filtering.

Even after the **Do8CCA(...) function** returns, the non-zero integer values do not all represent true EAs as defined by SSCC rules. Thus the next stage of the process is the filtering and this is implemented in a separate code module because the filtering can be based on anything. This code module, although separate from the main module, nevertheless still acts on the same global integer array. The SSCC filtering is done in two stages. The first is called **Perimeter Filtering** and is based on the idea previously described that any empty square that is part of a horizontal or vertical or combination-horizontal-and-vertical path of empty squares leading all the way to the rim of the chessboard is NOT part of an SSCC EA. See Figure 3 (a) for a pictorial representation of why this is the case. After Perimeter Filtering, the second stage is **Minimum EA Size Filtering**. This basically removes any EA that is below the minimum size (in terms of number of array elements) required. For SSCC, this minimum size is two array elements so the only EAs that get eliminated are those that have only a single element. But the filtering can remove EAs below any size for applications beyond SSCC.

The perimeter filtering first sets the integer values at the "4 corners" of the global integer classes array (g_iClassesArray) to 0. It then runs along each perimeter of the array; first the top, then the left side, then the right side, and finally the bottom, and sets each edge element to 0 plus elements that are horizontally and / or vertically connected to the perimeter element. Here is the overall pseudocode of the 2 stages of the filtering. The lines beginning with "//----" are merely comment lines that do not compile, only elucidate.

```
//////////////////////////////////////////////////////////////////////////////////
//---- Pseudocode for SSCC filtering
//---- The argument to this function is the minimum number of connected elements needed
//---- to consider an EA being valid. In the case of SSCC, this value is 2.
//----
void Filter8ConnectedClassesForSSCC( unsigned integer nMinimumClassPixelsNeeded )
{
   //---- Make a copy of g_iClassesArray, we will not change the copy
   DECLARE integer array[64] unchangedClassesArray;
   FOR integer i = 0 to (ARRAY_SIZE - 1) incrementing by 1
      SET unchangedClassesArray[i] = g_iClassesArray[i]
   NEXT FOR

   //---- PERIMETER FILTERING
   FOR integer i1 = 0 to (ARRAY_SIZE - 1) incrementing by 1        //---- ARRAY_SIZE is 64 for our 8x8 array
      integer iClassNum = unchangedClassesArray[i1]

      IF i1 indexes top-left, top-right, bottom-left, or bottom-right corner of 8x8 array
         SET g_iClassesArray[i1] = 0
      ELSE IF i1 indexes element along top row of 8x8 array
         IF iClassNum > 0
            SET g_iClassesArray[i1] = 0
            SET integer i2 = i1 + ARRAY_WIDTH                    //---- ARRAY_WIDTH is 8 for our 8x8 array
            WHILE i2 < ARRAY_SIZE
               integer iTest = unchangedClassesArray[i2]
               IF iTest > 0
                  SET g_iClassesArray[i2] = 0
                  //---- This function sets all horizontal / vertical connected elements to 0.
                  //---- Second argument is an enumerated value indicating start direction
                  //---- for finding connected elements.
                  CALL CheckXYConnectedElements( i2, eLeftDirection, unchangedClassesArray )
                  SET i2 = i2 + ARRAY_WIDTH
               ELSE
                  BREAK FROM WHILE LOOP
            END WHILE
         END IF
      ELSE IF i1 indexes element along leftmost column of 8x8 array
         IF iClassNum > 0
            SET g_iClassesArray[i1] = 0
            SET integer i2 = i1 + 1
            WHILE i2 does not index element on right perimeter of 8x8 array
               integer iTest = unchangedClassesArray[i2]
               IF iTest > 0
                  SET g_iClassesArray[i2] = 0
                  //---- This function sets all horizontal / vertical connected elements to 0.
                  //---- Second argument is an enumerated value indicating start direction
                  //---- for finding connected elements.
                  CALL CheckXYConnectedElements( i2, eUpDirection, unchangedClassesArray )
                  SET i2 = i2 + 1
               ELSE
                  BREAK FROM WHILE LOOP
```

```
            END WHILE
        END IF
    ELSE IF i1 indexes element along rightmost column of 8x8 array
        IF iClassNum > 0
            SET g_iClassesArray[i1] = 0
            SET integer i2 = i1 - 1
            WHILE i2 does not index element on left perimeter of 8x8 array
                integer iTest = unchangedClassesArray[i2]
                IF iTest > 0
                    SET g_iClassesArray[i2] = 0
                    //---- This function sets all horizontal / vertical connected elements to 0.
                    //---- Second argument is an enumerated value indicating start direction
                    //---- for finding connected elements.
                    CALL CheckXYConnectedElements( i2, eUpDirection, unchangedClassesArray )
                    SET i2 = i2 - 1
                ELSE
                    BREAK FROM WHILE LOOP
            END WHILE
        END IF
    ELSE IF i1 indexes element along bottom row of 8x8 array
        IF iClassNum > 0
            SET g_iClassesArray[i1] = 0
            SET integer i2 = i1 - ARRAY_WIDTH
            WHILE i2 > 0
                integer iTest = unchangedClassesArray[i2]
                IF iTest > 0
                    SET g_iClassesArray[i2] = 0
                    //---- This function sets all horizontal / vertical connected elements to 0.
                    //---- Second argument is an enumerated value indicating start direction
                    //---- for finding connected elements.
                    CALL CheckXYConnectedElements( i2, eLeftDirection, unchangedClassesArray )
                    SET i2 = i2 - ARRAY_WIDTH
                ELSE
                    BREAK FROM WHILE LOOP
            END WHILE
        END IF
    END IF
  NEXT FOR
  //---- END OF PERIMETER FILTERING

  //---- This call eliminates all EAs in g_iClassesArray that are below the minimum size.
  //---- For SSCC, this minimum size is 2.
  SET integer iNumOfClassesFiltered = CALL EliminateClassesBelowSize( nMinimumClassPixelsNeeded  )

  IF iNumOfClassesFiltered > 0
    FOR integer i = 0 to (ARRAY_SIZE - 1) incrementing by 1
        SET integer iClassNum = g_iClassesArray[i]
        SET integer iTestClassNum = 1
        SET integer iDecrement = 0
        WHILE (iTestClassNum < iClassNum  AND  iDecrement < iNumOfClassesFiltered )
            //---- g_ClassesFiltered is a set of class numbers that were filtered out
            //---- in the above CALL to EliminateClassesBelowSize(...)
            IF iTestClassNum found in g_ClassesFiltered
                SET iDecrement = iDecrement + 1
            END IF
            SET iTestClassNum = iTestClassNum + 1
        END WHILE
        SET iClassNum = iClassNum - iDecrement
        IF iClassNum > 1
            SET g_ClassesArray[i] = iClassNum
        END IF
    NEXT FOR
  END IF
}
```

The above function may call 2 other functions (the lines in bold above), and the pseudocode (with a preceding explanation) is given here of each of them. Here is the first one:

```
////////////////////////////////////////////////////////////////////////////////
//---- Pseudocode for checkXYZConnectedelements(...)
//---- SearchDirection is an enumeration specifying starting search direction in global classes array.
//---- SearchDirection can be any one of Left, Right, Up, Down
//---- Third argument is an copy of the global classes array, before any values changed by filtering.
```

```
//----
void checkXYConnectedElements( integer const iStartIndex,  SearchDirection eSD, integer[] unchangedClassesArray )
{
   DECLARE vector< pair<integer, Searchdirection> > IndexesVector
   DECLARE pair<integer, SearchDirection> nextPair
   SET nextPair.integer = iStartIndex
   SET nextPair.SearchDirection = eSD
   SET integer iTextIndex = -1
   SET SearchDirection searchDir = eSD
   SET integer iTest = -1

   DO
      IF searchDir is Left or Right
         IF searchDir is Left
            SET iTestIndex = nextPair.integer - 1
            SET iTest = 0
            WHILE iTestIndex does not index element on left perimeter of 8x8 array
               SET iTest = unchangedClassesArray[iTestIndex]
               IF iTest > 0
                  SET integer iTest2 = g_iClassesArray[iTestIndex]
                  IF iTest2 > 0
                     SET g_iClassesArray[iTestIndex] = 0
                     SET pair<integer, SearchDirection> tempPair1 = (iTestIndex, Left)
                     INSERT tempPair1 to back of IndexesVector
                     SET pair<integer, SearchDirection> tempPair2 = (iTestIndex, Up)
                     INSERT tempPair2 to back of IndexesVector
                     //---- While GOTO statements are generally frowned upon, it is used here for simplicity in avoiding function recursion,
                     //---- which is even more frowned upon since it limits space complexity to the size of the program call stack.
                     GOTO OuterLoopEnd
                  ELSE
                     iTestIndex = iTestIndex - 1
                  END IF
               ELSE
                  BREAK from WHILE
               END IF
            END WHILE
            SET searchDir = Right
         END IF
         IF searchDir is Right
            SET iTestIndex = nextPair.integer + 1
            IF IndexesVector size > 0
               REMOVE back element from IndexesVector
            END IF
            SET iTest = 0
            WHILE iTestIndex does not index element on right perimeter of 8x8 array
               SET iTest = unchangedClassesArray[iTestIndex]
               IF iTest > 0
                  SET integer iTest2 = g_iClassesArray[iTestIndex]
                  IF iTest2 > 0
                     SET g_iClassesArray[iTestIndex] = 0
                     SET pair<integer, SearchDirection> tempPair1 = (iTestIndex, Right)
                     INSERT tempPair1 to back of IndexesVector
                     SET pair<integer, SearchDirection> tempPair2 = (iTestIndex, Up)
                     INSERT tempPair2 to back of IndexesVector
                     //---- While GOTO statements are generally frowned upon, it is used here for simplicity in avoiding function recursion,
                     //---- which is even more frowned upon since it limits space complexity to the size of the program call stack.
                     GOTO OuterLoopEnd
                  ELSE
                     iTestIndex = iTestIndex + 1
                  END IF
               ELSE
                  BREAK from WHILE
               END IF
            END WHILE
         END IF
      ELSE IF searchDir is Up or Down
         IF searchDir is Up
            SET iTestIndex = nextPair.integer - ARRAY_WIDTH        //----ARRAY_WIDTH is 8 for our 8x8 array.
            SET iTest = 0
            WHILE iTestIndex > 0
               SET iTest = unchangedClassesArray[iTestIndex]
               IF iTest > 0
                  SET integer iTest2 = g_iClassesArray[iTestIndex]
                  IF iTest2 > 0
                     SET g_iClassesArray[iTestIndex] = 0
```

```
                     SET pair<integer, SearchDirection> tempPair1 = (iTestIndex, Up)
                     INSERT tempPair1 to back of IndexesVector
                     SET pair<integer, SearchDirection> tempPair2 = (iTestIndex, Left)
                     INSERT tempPair2 to back of IndexesVector
                     //---- While GOTO statements are generally frowned upon, it is used here for simplicity in avoiding function recursion,
                     //---- which is even more frowned upon since it limits space complexity to the size of the program call stack.
                     GOTO OuterLoopEnd
                  ELSE
                     iTestIndex = iTestIndex - ARRAY_WIDTH
                  END IF
               ELSE
                  BREAK from WHILE
               END IF
            END WHILE
            SET searchDir = Down
         END IF
         IF searchDir is Down
            SET iTestIndex = nextPair.integer + ARRAY_WIDTH
            IF IndexesVector size > 0
               REMOVE back element from IndexesVector
            END IF
            SET iTest = 0
            WHILE iTestIndex < ARRAY_SIZE                   //---- ARRAY_SIZE is 64 for our 8x8 array
               SET iTest = unchangedClassesArray[iTestIndex]
               IF iTest > 0
                  SET integer iTest2 = g_iClassesArray[iTestIndex]
                  IF iTest2 > 0
                     SET g_iClassesArray[iTestIndex] = 0
                     SET pair<integer, SearchDirection> tempPair1 = (iTestIndex, Down)
                     INSERT tempPair1 to back of IndexesVector
                     SET pair<integer, SearchDirection> tempPair2 = (iTestIndex, Left)
                     INSERT tempPair2 to back of IndexesVector
                     //---- While GOTO statements are generally frowned upon, it is used here for simplicity in avoiding function recursion,
                     //---- which is even more frowned upon since it limits space complexity to the size of the program call stack.
                     GOTO OuterLoopEnd
                  ELSE
                     iTestIndex = iTestIndex + ARRAY_WIDTH
                  END IF
               ELSE
                  BREAK from WHILE
               END IF
            END WHILE
         END IF
      END IF
OuterLoopEnd:
      IF IndexesVector size is 0
         BREAK from DO-WHILE
      SET nextPair = back element of IndexesVector
      SET searchDir = nextPair.SearchDirection
   WHILE IndexesVector size > 0
}
```

Here is the pseudocode for the second function called from within the very first function outlined above:

```
////////////////////////////////////////////////////////////////////////////////
//---- Pseudocode for eliminateClassesBelowSize
//---- Returns the number of classes that were filtered out because the EA was below minimum size.
//---- For SSCC, the minimum size is 2.
//----
unsigned integer eliminateClassesBelowSize( unsigned integer nMinSize )
{
   SET unsigned int iReturn = 0
   IF nMinSize > 1  OR  nMinSize > (ARRAY_SIZE / 2)
      RETURN iReturn
   //---- Clear the global set of class numbers that have been filtered out
   CLEAR g_ClassesFilteredSet
   DECLARE set<int> IndexesForChangeSet
   DECLARE set<int> IndexesCheckedSet

   //---- ConnectedElement is an enumeration indicating which array element relative to the current one.
   //---- It can be any one of TopLeft, TopCenter, TopRight, Left, Right, BottomLeft, BottomCenter, BottomRight
   DECLARE vector<pair<int, ConnectedElement>> TestIndexesVector
```

```
  FOR integer iIndex from 0 to (ARRAY_SIZE - 1), incrementing by 1
    IF IndexesCheckedSet contains iIndex
      NEXT FOR
    END IF
    INSERT iIndex into IndexesCheckedSet
    CLEAR IndexesForChangeSet
    SET unsigned integer nCounter = 0
    IF g_iClassesArray[iIndex] <= 0
      NEXT FOR
    nCounter = nCounter + 1
    INSERT iIndex into IndexesForChangeSet
    SET integer iTestIndex = iIndex - (ARRAY_WIDTH + 1)
    SET integer iOriginatingIndex = iIndex
    SET ConnectedElement nextConnectedElement = TopLeft

    DO
      IF nCounter > 1
        SET pair<int, ConnectedElement> nextPair = back element of TestIndexesVector
        iTestIndex = nextPair.integer
        nextConnectedElement = nextPair.ConnectedElement
      END IF
      SET integer iNextClassNumber = 0
      SET boolean bPopBack = true
NextTestIndex:
      IF nextConnectedElement is TopLeft
        SET iOriginatingIndex = iTestIndex + ARRAY_WIDTH + 1
        IF (iTestIndex >= 0  AND  0 < remainder of iOriginatingIndex / ARRAY_WIDTH )
          IF iTestIndex not found in IndexesCheckedSet
            SET iNextClassNumber = g_iClassesArray[iTestIndex]
            IF iNextClassNumber equal to iStartClassNumber
              SET nCounter = nCounter + 1
              INSERT iTestIndex into IndexesCheckedSet
              INSERT itestIndex into IndexesForChangeSet
              SET pair<integer, ConnectedElement> nextPair = (iTestIndex + 1, TopCenter)
              INSERT nextPair to back of TestIndexesVector
              SET bPopBack = false
              SET iTestIndex = iTestIndex - (ARRAY_WIDTH + 1)
              SET nextConnectedElement = TopLeft
              //---- While GOTO statements are generally frowned upon, it is used here for simplicity in avoiding function recursion,
              //---- which is even more frowned upon since it limits space complexity to the size of the program call stack.
              GOTO NextTestIndex
            END IF
          END IF
        END IF
        SET iTestIndex = iTestIndex + 1
      END IF

      IF nextConnectedElement is TopCenter
        SET iOriginatingIndex = iTestIndex + ARRAY_WIDTH
        IF iTestIndex >= 0
          IF iTestIndex not found in IndexesCheckedSet
            SET iNextClassNumber = g_iClassesArray[iTestIndex]
            IF iNextClassNumber equal to iStartClassNumber
              SET nCounter = nCounter + 1
              INSERT iTestIndex into IndexesCheckedSet
              INSERT itestIndex into IndexesForChangeSet
              SET pair<integer, ConnectedElement> nextPair = (iTestIndex + 1, TopRight)
              INSERT nextPair to back of TestIndexesVector
              SET bPopBack = false
              SET iTestIndex = iTestIndex - (ARRAY_WIDTH + 1)
              SET nextConnectedElement = TopLeft
              //---- While GOTO statements are generally frowned upon, it is used here for simplicity in avoiding function recursion,
              //---- which is even more frowned upon since it limits space complexity to the size of the program call stack.
              GOTO NextTestIndex
            END IF
          END IF
        END IF
        SET iTestIndex = iTestIndex + 1
      END IF

      IF nextConnectedElement is TopRight
        SET iOriginatingIndex = iTestIndex + ARRAY_WIDTH - 1
        IF (iTestIndex >= 0  AND  (ARRAY_WIDTH - 1) > remainder of iOriginatingIndex / ARRAY_WIDTH )
          IF iTestIndex not found in IndexesCheckedSet
            SET iNextClassNumber = g_iClassesArray[iTestIndex]
```

```
            IF iNextClassNumber equal to iStartClassNumber
                SET nCounter = nCounter + 1
                INSERT iTestIndex into IndexesCheckedSet
                INSERT itestIndex into IndexesForChangeSet
                SET pair<integer, ConnectedElement> nextPair = (iTestIndex + ARRAY_WIDTH - 2, Left)
                INSERT nextPair to back of TestIndexesVector
                SET bPopBack = false
                SET iTestIndex = iTestIndex - (ARRAY_WIDTH + 1)
                SET nextConnectedElement = TopLeft
                //---- While GOTO statements are generally frowned upon, it is used here for simplicity in avoiding function recursion,
                //---- which is even more frowned upon since it limits space complexity to the size of the program call stack.
                GOTO NextTestIndex
            END IF
        END IF
    END IF
    SET iTestIndex = iTestIndex + ARRAY_WIDTH - 2
END IF

IF nextConnectedElement is Left
    SET iOriginatingIndex = iTestIndex + 1
    IF (iTestIndex >= 0  AND  0 < remainder of iOriginatingIndex / ARRAY_WIDTH )
        IF iTestIndex not found in IndexesCheckedSet
            SET iNextClassNumber = g_iClassesArray[iTestIndex]
            IF iNextClassNumber equal to iStartClassNumber
                SET nCounter = nCounter + 1
                INSERT iTestIndex into IndexesCheckedSet
                INSERT itestIndex into IndexesForChangeSet
                SET pair<integer, ConnectedElement> nextPair = (iTestIndex + 2, Right)
                INSERT nextPair to back of TestIndexesVector
                SET bPopBack = false
                SET iTestIndex = iTestIndex - (ARRAY_WIDTH + 1)
                SET nextConnectedElement = TopLeft
                //---- While GOTO statements are generally frowned upon, it is used here for simplicity in avoiding function recursion,
                //---- which is even more frowned upon since it limits space complexity to the size of the program call stack.
                GOTO NextTestIndex
            END IF
        END IF
    END IF
    SET iTestIndex = iTestIndex + 2
END IF

IF nextConnectedElement is Right
    SET iOriginatingIndex = iTestIndex - 1
    IF (iTestIndex < ARRAY_SIZE  AND  (ARRAY_WIDTH - 1) > remainder of iOriginatingIndex / ARRAY_WIDTH )
        IF iTestIndex not found in IndexesCheckedSet
            SET iNextClassNumber = g_iClassesArray[iTestIndex]
            IF iNextClassNumber equal to iStartClassNumber
                SET nCounter = nCounter + 1
                INSERT iTestIndex into IndexesCheckedSet
                INSERT itestIndex into IndexesForChangeSet
                SET pair<integer, ConnectedElement> nextPair = (iTestIndex + ARRAY_WIDTH - 2, BottomLeft)
                INSERT nextPair to back of TestIndexesVector
                SET bPopBack = false
                SET iTestIndex = iTestIndex - (ARRAY_WIDTH + 1)
                SET nextConnectedElement = TopLeft
                //---- While GOTO statements are generally frowned upon, it is used here for simplicity in avoiding function recursion,
                //---- which is even more frowned upon since it limits space complexity to the size of the program call stack.
                GOTO NextTestIndex
            END IF
        END IF
    END IF
    SET iTestIndex = iTestIndex + ARRAY_WIDTH - 2
END IF

IF nextConnectedElement is BottomLeft
    SET iOriginatingIndex = iTestIndex - (ARRAY_WIDTH - 1)
    IF (iTestIndex < ARRAY_SIZE  AND  0 < remainder of iOriginatingIndex / ARRAY_WIDTH )
        IF iTestIndex not found in IndexesCheckedSet
            SET iNextClassNumber = g_iClassesArray[iTestIndex]
            IF iNextClassNumber equal to iStartClassNumber
                SET nCounter = nCounter + 1
                INSERT iTestIndex into IndexesCheckedSet
                INSERT itestIndex into IndexesForChangeSet
                SET pair<integer, ConnectedElement> nextPair = (iTestIndex + 1, BottomCenter)
                INSERT nextPair to back of TestIndexesVector
```

```
                    SET bPopBack = false
                    SET iTestIndex = iTestIndex - (ARRAY_WIDTH + 1)
                    SET nextConnectedElement = TopLeft
                    //---- While GOTO statements are generally frowned upon, it is used here for simplicity in avoiding function recursion,
                    //---- which is even more frowned upon since it limits space complexity to the size of the program call stack.
                    GOTO NextTestIndex
                END IF
            END IF
        END IF
        SET iTestIndex = iTestIndex + 1
    END IF

    IF nextConnectedElement is BottomCenter
        SET iOriginatingIndex = iTestIndex - ARRAY_WIDTH
        IF (iTestIndex < ARRAY_SIZE )
            IF iTestIndex not found in IndexesCheckedSet
                SET iNextClassNumber = g_iClassesArray[iTestIndex]
                IF iNextClassNumber equal to iStartClassNumber
                    SET nCounter = nCounter + 1
                    INSERT iTestIndex into IndexesCheckedSet
                    INSERT itestIndex into IndexesForChangeSet
                    SET pair<integer, ConnectedElement> nextPair = (iTestIndex + 1, BottomRight)
                    INSERT nextPair to back of TestIndexesVector
                    SET bPopBack = false
                    SET iTestIndex = iTestIndex - (ARRAY_WIDTH + 1)
                    SET nextConnectedElement = TopLeft
                    //---- While GOTO statements are generally frowned upon, it is used here for simplicity in avoiding function recursion,
                    //---- which is even more frowned upon since it limits space complexity to the size of the program call stack.
                    GOTO NextTestIndex
                END IF
            END IF
        END IF
        SET iTestIndex = iTestIndex + 1
    END IF

    IF nextConnectedElement is BottomRight
        SET iOriginatingIndex = iTestIndex - (ARRAY_WIDTH + 1)
        IF (iTestIndex < ARRAY_SIZE  AND  (ARRAY_WIDTH - 1) > remainder of iOriginatingIndex / ARRAY_WIDTH )
            IF iTestIndex not found in IndexesCheckedSet
                SET iNextClassNumber = g_iClassesArray[iTestIndex]
                IF iNextClassNumber equal to iStartClassNumber
                    SET nCounter = nCounter + 1
                    INSERT iTestIndex into IndexesCheckedSet
                    INSERT itestIndex into IndexesForChangeSet
                    SET iTestIndex = iTestIndex - (ARRAY_WIDTH + 1)
                    SET nextConnectedElement = TopLeft
                     //---- While GOTO statements are generally frowned upon, it is used here for simplicity in avoiding function recursion,
                    //---- which is even more frowned upon since it limits space complexity to the size of the program call stack.
                    GOTO NextTestIndex
                END IF
            END IF
        END IF
    END IF

    IF (bPopBack is true  AND  TestIndexesVector is not empty)
        REMOVE back element from TestIndexesVector
    END IF
    SET bPopBack = true
  WHILE TestIndexesVector is not empty

  IF (nCounter > 0  AND  nCounter < nMinSize )
    INSERT iStartClassNumber into g_classesFilteredSet
    FOREACH integer iIndexForChange IN IndexesForChangeSet
        g_iClassesArray[iIndexForChange] = 0
    END FOREACH
    SET iReturn = iReturn + 1
  END IF
NEXT FOR
RETURN iReturn
}
```

Once the Do8CCA(...) and Filter8ConnectedClassesForSSCC(...) functions have completed, it is possible that class numbering could be off. That is, if the Do8CCA(...) function modified the global classes array and left it containing, say, 4 separate EAs numbered 1 to 4, and then the filtering removed classes numbered 2 and 3, the global array would be left with two EAs, but numbered 1 and 4. To correct this, the Do8CCA(...) function is called a second time, and this always corrects the numbering. For the purposes of SSCC and detecting valid chains, this is not necessary. It is done just for a sense of correctness, but if placed in an actual SSCC game engine and performance needed to be optimized, this second call to Do8CCA(...) could be removed.

After these functions have finished executing, the g_iClassesArray contains values > 0 only for the correct elements that are part of an SSCC EA. This has been validated by extensive testing over many hundreds of random positions, leaving the remote possibility that there could be one or more "corner cases" in which the above code produces an erronous result. There is no known mathematical proof that the above code methods are infallible, but if such corner cases do indeed exist, they must be very rare and special indeed.

Finally comes the meat of the algorithm and the part that is, at least for now, unique and the point of this article: a single function (with two helper functions) that takes what we have so far -- the global class array with any EAs, and the index into that array of the last-moved-piece location -- and returns a boolean true or false that correctly tells us whether there are two paths from the last-piece-moved location to one and the same EA, where the two paths do not share a common element.

Before listing that function's pseudocode, a few things need to be explained. The various paths that are found will each be represented by a standard programming data structure called a list. The basic property of each node of a list is that, in addition to containing a piece of data (in this case, an integer index into the global classes array, which is equivalent to the node's location on the chessboard as shown in the table of Figure 7), the node also points to the next node in the list. Thus once you have any node in the list, you can progress forward through the list from there. This is perfect for our notion of a "valid" path of board squares that lead from the last-piece-moved location to an EA. The function will be creating many such lists, some valid and some not valid (in SSCC terms), and to keep track of them all, we place them in a vector (a resizeable array). Thus we have the following global data structure for tracking paths to an EA:

**vector<list<integer>> g_EAPathListVector**

This is read as a vector containing from 0 to many lists, each list containing 0 to many nodes of data type integer (and since it is a list node, you know that each node contains something that points to the next node in the list). The name of this global data structure is g_EAPathListVector. That is how the function will refer to it, by its name. Another structure that is needed is a multimap that tracks, for each unique EA, the path lists that have been found leading to that EA from the last-piece-moved square. A multimap is like a dictionary, where for each key you can have multiple values. Thus we have the following global data structure for tracking this:

**multimap<integer, integer> g_EAPathClassMultiMap**

Secondly, since each list will need to be composed of integer index values that have not been used by another list currently in process (because the paths must have no common element), we need to keep track of indexes that have been used during each one of the many cycles of the process. Since each index is really a way to refer to a square on the chessboard, we know that there can only be ARRAY_SIZE of them, and for our 8x8 chessboard, this means 64 of them. So we have another global integer array,

**integer[ARRAY_SIZE] g_iUsedBoardSquares**

to keep track of which indexes (BoardSquares) we have already used. With all that explained, here now is the pseudocode for the function:

```
////////////////////////////////////////////////////////////////////////////////////////////
//---- Pseudocode for found2PathsToClassFromIndex(...)
//---- Returns true if 2 unique paths are found from the board square given by
//---- the crLastPieceMovedIndex argument to the same EA (class) number.
//---- All arguments having name prefixed by "r" are passed by reference, meaning that
//---- any changes to that object within the function are changing the caller's object.
//---- All arguments having name prefixed by "cr" are passed by const reference, meaning
//---- that they cannot be changed in any way by the function.
//---- The global g_iClassesArray is the same global integer array as in other functions.
//
```

```
boolean found2PathsToClassFromIndex
  ( boolean rbAllDone,
    integer iPath,
    integer iLevel,
    unsigned integer crLastPieceMovedIndex )
{
    CLEAR g_EAPathListVector
    CLEAR g_EAPathClassMultiMap
    FOR EACH integer i in g_iUsedBoardSquares
        SET i = 0
    END FOR EACH

    DECLARE boolean bFoundPathtoClass
    DECLARE integer iIndex, iTryIndex, iTopLeftIndex, iTopCenterIndex, iTopRightIndex,
                iLeftIndex, iRightIndex, iBottomLeftIndex, iBottomCenterIndex,
                iBottomRightIndex, iLastForIndex
    DECLARE list<integer> PathList
    DECLARE set<integer> VisitedSquaresSet
    DECLARE vector<int> StartForLoopIndexesVector;

    SET bFoundPathToClass = false
    SET iLastForIndex = 0
    INSERT iLastForIndex to back of StartForLoopIndexesVector
    SET iIndex = crLastPieceMovedIndex

    IF iLevel is 1
        SET iTopLeftIndex = iIndex - (ARRAY_WIDTH + 1)          //---- ARRAY_WIDTH is 8 for our 8x8 array
        SET iTopCenterIndex = iIndex - (ARRAY_WIDTH)
        sET iTopRightIndex = iIndex - (ARRAY_WIDTH - 1)
        SET iLeftIndex = iIndex - 1
        SET iRightIndex - iIndex + 1
        SET iBottomLeftIndex = iIndex + (ARRAY_WIDTH - 1)
        SET iBottomCenterIndex = iIndex + ARRAY_WIDTH
        SET iBottomRightIndex = iIndex + (ARRAY_WIDTH + 1)

        SET g_iClassesArray[iIndex] = 1
        IF iTopLeftIndex >= 0
            SET g_iClassesArray[iTopLeftIndex] = 1
        END IF
        IF iTopCenterIndex >= 0
            SET g_iClassesArray[iTopCenterIndex] = 1
        END IF
        IF iTopRightIndex >= 0
            SET g_iClassesArray[iTopRightIndex] = 1
        END IF
        IF iLeftIndex >= 0
            SET g_iClassesArray[iLeftIndex] = 1
        END IF
        IF iRightIndex < ARRAY_SIZE                              //---- ARRAY_SIZE is 64 for our 8x8 array
            SET g_iClassesArray[iRightIndex] = 1
        END IF
        IF iBottomLeftIndex < ARRAY_SIZE
            SET g_iClassesArray[iBottomLeftIndex] = 1
        END IF
        IF iBottomCenterIndex < ARRAY_SIZE
            SET g_iClassesArray[iBottomCenterIndex] = 1
        END IF
        IF iBottomRightIndex < ARRAY_SIZE
            SET g_iClassesArray[iBottomRightIndex] = 1
        END IF
    END IF

Recurse:
    SET iTryIndex = 0
    SET iTopLeftIndex = iIndex - (ARRAY_WIDTH + 1)
    SET iTopCenterIndex = iIndex - (ARRAY_WIDTH)
    sET iTopRightIndex = iIndex - (ARRAY_WIDTH - 1)
    SET iLeftIndex = iIndex - 1
    SET iRightIndex - iIndex + 1
    SET iBottomLeftIndex = iIndex + (ARRAY_WIDTH - 1)
    SET iBottomCenterIndex = iIndex + ARRAY_WIDTH
    SET iBottomRightIndex = iIndex + (ARRAY_WIDTH + 1)

    //---- NUM_SURROUNDING_ELEMENTS is 8 for 2D, and would be 26 for 3D
    FOR integer i from back element of StartForLoopIndexesVector to NUM_SURROUNDING_ELEMENTS incrementing by 1
```

```
SET boolean bValidIndex = false
IF iLevel is 1
   SET iPath = number of elements in g_EAPathListVector
END IF
IF i is 0
   SET iTryIndex = iTopLeftIndex
   IF (iTryIndex >= 0  AND  0 < remainder of iIndex / ARRAY_WIDTH)
      SET bValidIndex = true
   END IF
ELSE IF i is 1
   SET iTryIndex = iTopCenterIndex
   IF iTryIndex >= 0
      SET bValidIndex = true
   END IF
ELSE IF i is 2
   SET iTryIndex = iTopRightIndex
   IF (iTryIndex >= 0  AND  (ARRAY_WIDTH - 1) > remainder of iIndex / ARRAY_WIDTH)
      SET bValidIndex = true
   END IF
ELSE IF i is 3
   SET iTryIndex = iLeftIndex
   IF (iTryIndex >= 0  AND  0 < remainder of iIndex / ARRAY_WIDTH)
      SET bValidIndex = true
   END IF
ELSE IF i is 4
   SET iTryIndex = iRightIndex
   IF (iTryIndex < ARRAY_SIZE  AND  (ARRAY_WIDTH - 1) > remainder of iIndex / ARRAY_WIDTH)
      SET bValidIndex = true
   END IF
ELSE IF i is 5
   SET iTryIndex = iBottomLeftIndex
   IF (iTryIndex < ARRAY_SIZE  AND  0 < remainder of iIndex / ARRAY_WIDTH)
      SET bValidIndex = true
   END IF
ELSE IF i is 6
   SET iTryIndex = iBottomCenterIndex
   IF iTryIndex < ARRAY_SIZE
      SET bValidIndex = true
   END IF
ELSE IF i is 7
   SET iTryIndex = iBottomRightIndex
   IF (iTryIndex < ARRAY_SIZE  AND  (ARRAY_WIDTH - 1) > remainder of iIndex / ARRAY_WIDTH)
      SET bValidIndex = true
   END IF
ELSE
   //---- TODO: add more clauses here to handle 3D
END IF-ELSE

IF bValidIndex is false
   NEXT FOR
ELSE IF (iLevel > 1   AND   g_iUsedBoardSquares[iTryIndex] is 1)
   NEXT FOR
ELSE IF iTryIndex is in VisitedSquaresSet
   NEXT FOR
END IF-ELSE

SET integer iClass = g_iClassesArray[iTryIndex]
IF iClass > 0
   NEXT FOR
ELSE IF g_pieceArray[iTryIndex] is not empty string
   //---- (g_pieceArray is the array of piece strings from parsing the SSCC FEN string)
   INSERT iTryIndex to back of PathList
   SET set<int> AdjacentClasses = CALL isAdjacentToEA( iTryIndex )
   IF AdjacentClasses is not empty
      SET bFoundPathToClass = true
      FOR EACH iClass in AdjacentClasses
         INSERT PathList to back of g_EAPathListVector
         SET pair<integer, integer> validClassElement = (iPath, iClass)
         INSERT validClassElement into g_EAPathClassMultiMap
         SET boolean b = CALL areTwoUniquePathsToClass(crLastPieceMovedIndex, iClass)
         IF b is true
            SET rbAllDone = true
            RETURN bFoundPathToClass
         ELSE
            SET iPath = iPath + 1
```

```
                END IF-ELSE
            END FOR EACH
            REMOVE back element from PathList
            NEXT FOR
        END IF
        INSERT iIndex into VisitedSquaresSet
        SET iLevel = iLevel + 1
        IF StartForLoopIndexesVector is not empty
            REMOVE back element from StartForLoopIndexesVector
        END IF
        INSERT (i + 1) to back of StartForLoopIndexesVector
        INSERT 0 to back of StartForLoopIndexesVector
        SET iIndex = iTryIndex
        SET bFoundPathToClass = false
        //---- While GOTO statements are generally frowned upon, it is used here for simplicity in avoiding function recursion,
        //---- which is even more frowned upon since it limits space complexity to the size of the program stack.
        GOTO Recurse
    END IF-ELSE
  NEXT FOR
  IF StartForLoopIndexesVector is not empty
    REMOVE back element from StartForLoopIndexesVector
  END IF
  IF iLevel > 1
    REMOVE back element from PathList
    REMOVE iIndex from VisitedSquaresSet
    IF bFoundPathToClass
        SET iPath = g_EAPathListVector size
    END IF
    SET iLevel = iLevel + 1
    IF (rbAllDone is false  AND  StartForLoopIndexesVector is not empty )
        IF PathList is not empty
            SET iIndex = back element of PathList
        ELSE
            SET iIndex = crLastPieceMovedIndex
        END IF-ELSE
        SET bFoundPathToClass = false
         //---- While GOTO statements are generally frowned upon, it is used here for simplicity in avoiding function recursion,
        //---- which is even more frowned upon since it limits space complexity to the size of the program stack.
       GOTO Recurse
    END IF
  END IF
  RETURN bFoundPathToClass
}
```

You can see that the above pseudocode calls into 2 other functions that haven't been listed yet. Here is pseudocode for the first of those functions:

```
//////////////////////////////////////////////////////////////////////////////
//---- Pseudocode for isAdjacentToEA(integer)
//---- Returns a set<integer> of all classes that are adjacent to the board square
//---- indexed by the iBoardInde parameter.
set<integer> isAdjacentToEA(integer iBoardIndex)
{
  DECLARE set<integer> ReturnSet
  DECLARE integer iTryIndex

  WHILE( true )
    //---- TopLeft from iBoardIndex
    SET iTryIndex = iBoardIndex - (ARRAY_WIDTH + 1)
    IF( iTryIndex >= 0  AND  0 < remainder of iBoardIndex / ARRAY_WIDTH)
        SET integer iClass = g_iClassesArray[iTryIndex]
        IF iClass > 0
            INSERT iClass into ReturnSet
        END IF
    END IF

    //---- TopCenter from iBoardIndex
    SET iTryIndex = iBoardIndex - ARRAY_WIDTH
    IF( iTryIndex >= 0 )
        SET integer iClass = g_iClassesArray[iTryIndex]
        IF iClass > 0
            INSERT iClass into ReturnSet
```

```
        END IF
    END IF

    //---- TopRight from iBoardIndex
    SET iTryIndex = iBoardIndex - (ARRAY_WIDTH - 1)
    IF( iTryIndex >= 0  AND  (ARRAY_WIDTH - 1) < remainder of iBoardIndex / ARRAY_WIDTH)
        SET integer iClass = g_iClassesArray[iTryIndex]
        IF iClass > 0
            INSERT iClass into ReturnSet
        END IF
    END IF

    //---- Left from iBoardIndex
    SET iTryIndex = iBoardIndex - 1
    IF( iTryIndex >= 0  AND  0 < remainder of iBoardIndex / ARRAY_WIDTH)
        SET integer iClass = g_iClassesArray[iTryIndex]
        IF iClass > 0
            INSERT iClass into ReturnSet
        END IF
    END IF

    //---- Right from iBoardIndex
    SET iTryIndex = iBoardIndex + 1
    IF( iTryIndex < ARRAY_SIZE  AND  (ARRAY_WIDTH - 1) > remainder of iBoardIndex / ARRAY_WIDTH)
        SET integer iClass = g_iClassesArray[iTryIndex]
        IF iClass > 0
            INSERT iClass into ReturnSet
        END IF
    END IF

    //---- BottomLeft from iBoardIndex
    SET iTryIndex = iBoardIndex + (ARRAY_WIDTH - 1)
    IF( iTryIndex < ARRAY_SIZE  AND  0 < remainder of iBoardIndex / ARRAY_WIDTH)
        SET integer iClass = g_iClassesArray[iTryIndex]
        IF iClass > 0
            INSERT iClass into ReturnSet
        END IF
    END IF

    //---- BottomCenter from iBoardIndex
    SET iTryIndex = iBoardIndex + ARRAY_WIDTH
    IF( iTryIndex < ARRAY_SIZE )
        SET integer iClass = g_iClassesArray[iTryIndex]
        IF iClass > 0
            INSERT iClass into ReturnSet
        END IF
    END IF

    //---- BottomRight from iBoardIndex
    SET iTryIndex = iBoardIndex + ARRAY_WIDTH + 1
    IF( iTryIndex < ARRAY_SIZE  AND  (ARRAY_WIDTH - 1) > remainder of iBoardIndex / ARRAY_WIDTH)
        SET integer iClass = g_iClassesArray[iTryIndex]
        IF iClass > 0
            INSERT iClass into ReturnSet
        END IF
    END IF

    BREAK
  END WHILE
  RETURN ReturnSet
}
```

The above function is fairly simple and straightforward, but the second function that is called from found2PathsToClassFromIndex(...) is a much different story. The authors wish to make clear that the design of this function, and the overall algorithm design of which it is a key part, may very well prove to be very inelegant and unoptimal. The authors only presents it as something that works for every corner case that has been encountered so far (which is a significant number!) and encourages other developers / programmers to work towards more elegant and optimized designs of their own for this most challenging problem. Here now is the pseudocode for the final piece of the algorithm:

```
////////////////////////////////////////////////////////////////////////////////
//---- Pseudocode for areTwoUniquePathsToClass( integer, integer )
//---- Returns true if two unique paths are found from the element in g_ipClassesArray
//---- indexed by iFromIndex leading to the class indicated by iClassToCount.
//---- Unique means the two paths must not share any common element.
//---- Otherwise returns false.
//
bool areTwoUniquePathsToClass( integer iFromIndex, integer iClassToCount )
{
  //---- These next two variables are declared as static, which in C++ means they are not
  //---- discarded when the function exits, but are kept in memory so that when
  //---- returning to this function, they are available and in the state they were
  //---- when last exiting the function. But unlike a global variable, they are
  //---- not available or accessible outside this function.
  DECLARE static vector<integer> ListIndexVector
  DECLARE static map<integer, vector<integer>> ClassToListIndexVectorsMap

  IF iClassToCount > 0

    IF ClassToListIndexVectorsMap contains iClasstoCount
      ListIndexVector = value in ClassToListIndexVectorsMap for key iClassToCount
    ELSE
      CLEAR ListIndexVector
    END IF-ELSE

    SET unsigned integer jLast = 1

    //---- g_EAPathClassMultiMap is a global multimap<integer, integer> which must be
    //---- cleared in main() before calling into found2PathsToClassFromIndex(...)
    //---- which is the caller of this function.
    IF g_EAPathClassMultiMap size is 1
      CLEAR ListIndexVector
      CLEAR ClasstoListIndexVectorsMap
      SET jLast = 1
    END IF

    //---- Before calling into this function from found2PathsToClassFromIndex(...),
    //---- an integer pair representing <Path #, Class#> was inserted into g_EAPathClassMultiMap.
    SET pair<integer, integer> nextPair = g_EAPathClassMultiMap back element
    SET integer iClass = nextPair second element
    IF iClass is equal to iClassToCount
      SET integer iIndex = nextPair first element
      INSERT iIndex to back of ListIndexVector
    END IF

    IF iClassCount > ClassToListIndexVectorsMap size
      //---- iClassToCount is being handled for the first time
      SET pair<integer, vector<integer>> tempPair = <iClassToCount, ListIndexVector>
      INSERT tempPair to back of ClassToListIndexVectorsMap
    ELSE
      //---- ListIndexVector is copied back into its slot in ClassToListIndexVectorsMap
      //---- because it may have been modified.
      INSERT ListIndexVector into ClassToListIndexVectorsMap for key value iClassToCount
    END IF-ELSE

    IF (g_EAPathClassMultiMap size is 1   OR   g_EAPathListVector size is 1 )
      RETURN false
    END IF

    SET unsigned integer j = 0
    IF ListIndexVector size > 0
      FOR integer i from 0 to (ListIndexVector size - 1) incrementing by 1
        SET integer iFirstIndex = ListIndexVector's[i]
        //---- The following assignment is done by reference, not by copy.
        //---- This makes for a significant improvement in speed.
        SET list<integer> rFirstPath = g_EAPathListVector[iFirstIndex]
        IF jLast is 0
          SET j = jLast
        ELSE
          SET j = i + 1
        END IF-ELSE
        IF j >= size of ListIndexVector
          SET j = i + 1
        END IF
```

```
            IF i < j
                WHILE j < size of ListIndexVector
                    SET integer iSecondIndex = ListIndexVector[j]
                    //---- The following assignment is done by reference, not by copy.
                    //---- This makes for a significant improvement in speed.
                    SET list<integer> rSecondPath = g_EAPathListVector[iSecondIndex]

                    SET boolean bMatchFound = false
                    FOR integer k from 0 to (rFirstPath size - 1)
                        SET integer iFirst = kth element in rPathList
                        SET bMatchFound = (iFirst found in rSecondList)
                        IF bMatchFound is true
                            BREAK from FOR loop
                    NEXT FOR
                    IF bMatchFound is false
                        //---- 2 unique paths have been found
                        //---- Code could be added here to log or output to screen
                        //---- the nodes of each unique path.
                        RETURN true
                    END IF
                    SET j = j + 1
                END WHILE
            END IF
        NEXT FOR
        SET jLast = j
    END IF

  END IF
  RETURN false
}
```

## APPENDIX B: FURTHER EXAMPLES AND OUTPUT SCREEN CAPTURES

An informative example discovered during testing of the algorithm is as shown in Figure 8 and then Figure 9.

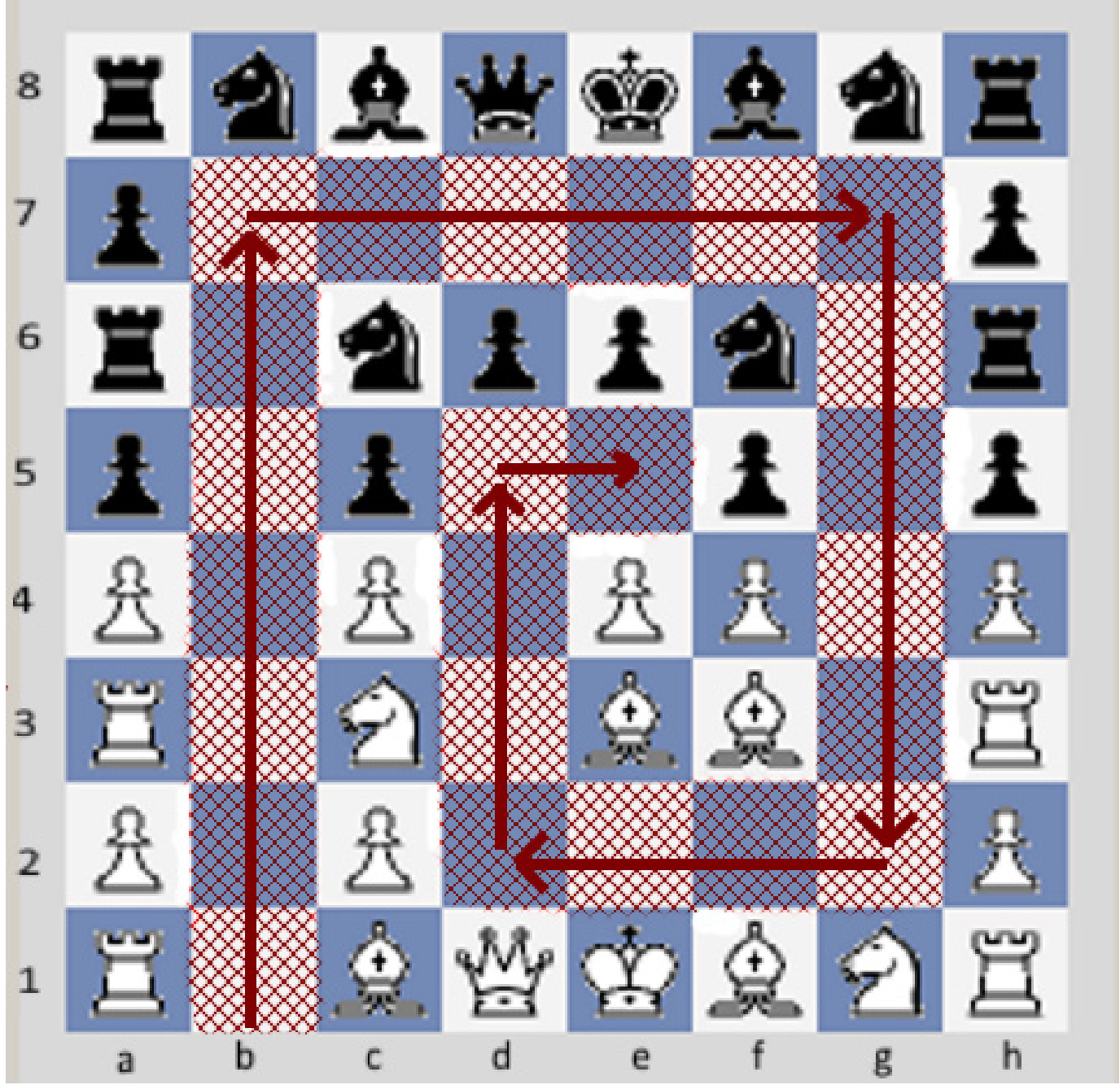

**Figure 8:** The example test case.

The position in Figure 8 is illegal in both chess and SSCC (too many pieces) but is used here to test the part of the algorithm that eliminates empty squares that are either directly on the rim of the board or are vertically or horizontally connected to such an empty rim square. There is a 'spiral' formed of empty board squares that has its beginning on the b1 square, which is on the rim, and ends up at the e5 square. For the purposes of this test, the e4 square was designated as the square to which the last piece to move moved but the results should be the same for any of the occupied squares so designated. The algorithm should initially assume that all these empty squares are connected and classify them as an EA with group identification '1'.

Then the 'perimeter filtering' portion of the algorithm should eliminate *all* of the empty squares, meaning there are no valid EA squares anywhere on the board and should thus bypass any further calculations. This is because there are *no* valid SSCC chains anywhere in this position. Here are the results that were written to screen by the program.

```
Board Position From FEN String:
        r       n       b       q       k       b       n       r
        p       -       -       -       -       -       -       p
        r       -       n       p       p       n       -       r
        p       -       p       -       -       p       -       p
        P       -       P       -       P       P       -       P
        R       -       N       -       B       B       -       R
        P       -       P       -       -       -       -       P
        R       -       B       Q       K       B       N       R
```

```
Classes After CCA:

        -       -       -       -       -       -       -       -
        -       1       1       1       1       1       1       -
        -       1       -       -       -       -       1       -
        -       1       -       1       1       -       1       -
        -       1       -       1       -       -       1       -
        -       1       -       1       -       -       1       -
        -       1       -       1       1       1       1       -
        -       1       -       -       -       -       -       -
(the algorithm correctly detected the spiral pattern, all connected as EA group #1)


Classes After CCA and Perimeter Filtering:
        -       -       -       -       -       -       -       -
        -       -       -       -       -       -       -       -
        -       -       -       -       -       -       -       -
        -       -       -       -       -       -       -       -
        -       -       -       -       -       -       -       -
        -       -       -       -       -       -       -       -
        -       -       -       -       -       -       -       -
        -       -       -       -       -       -       -       -
(the algorithm correctly eliminated the entirety of EA group #1)

Classes After CCA And Filtering:
        -       -       -       -       -       -       -       -
        -       -       -       -       -       -       -       -
        -       -       -       -       -       -       -       -
        -       -       -       -       -       -       -       -
        -       -       -       -       -       -       -       -
        -       -       -       -       -       -       -       -
        -       -       -       -       -       -       -       -
        -       -       -       -       -       -       -       -

Classes After CCA And Filtering -- Corrected:
        -       -       -       -       -       -       -       -
        -       -       -       -       -       -       -       -
        -       -       -       -       -       -       -       -
        -       -       -       -       -       -       -       -
        -       -       -       -       -       -       -       -
        -       -       -       -       -       -       -       -
        -       -       -       -       -       -       -       -
        -       -       -       -       -       -       -       -
A VALID CHAIN WAS ***NOT*** FOUND... Try another position?   y/n :
```

Since by the 3rd pass above there were no EAs left, the 4th pass where the output begins "Classes After CCA And Filtering -- Corrected" has actually done nothing. Only if there were multiple EAs and there was a gap in the numbering of them after the 3rd pass would the 4th pass actually be doing any correction, and as previously mentioned, this correction is not a requirement for the algorithm to work and is only included for a sense of "completeness". Now we move on to Figure 9 that required the algorithm to be adapted in order to handle.

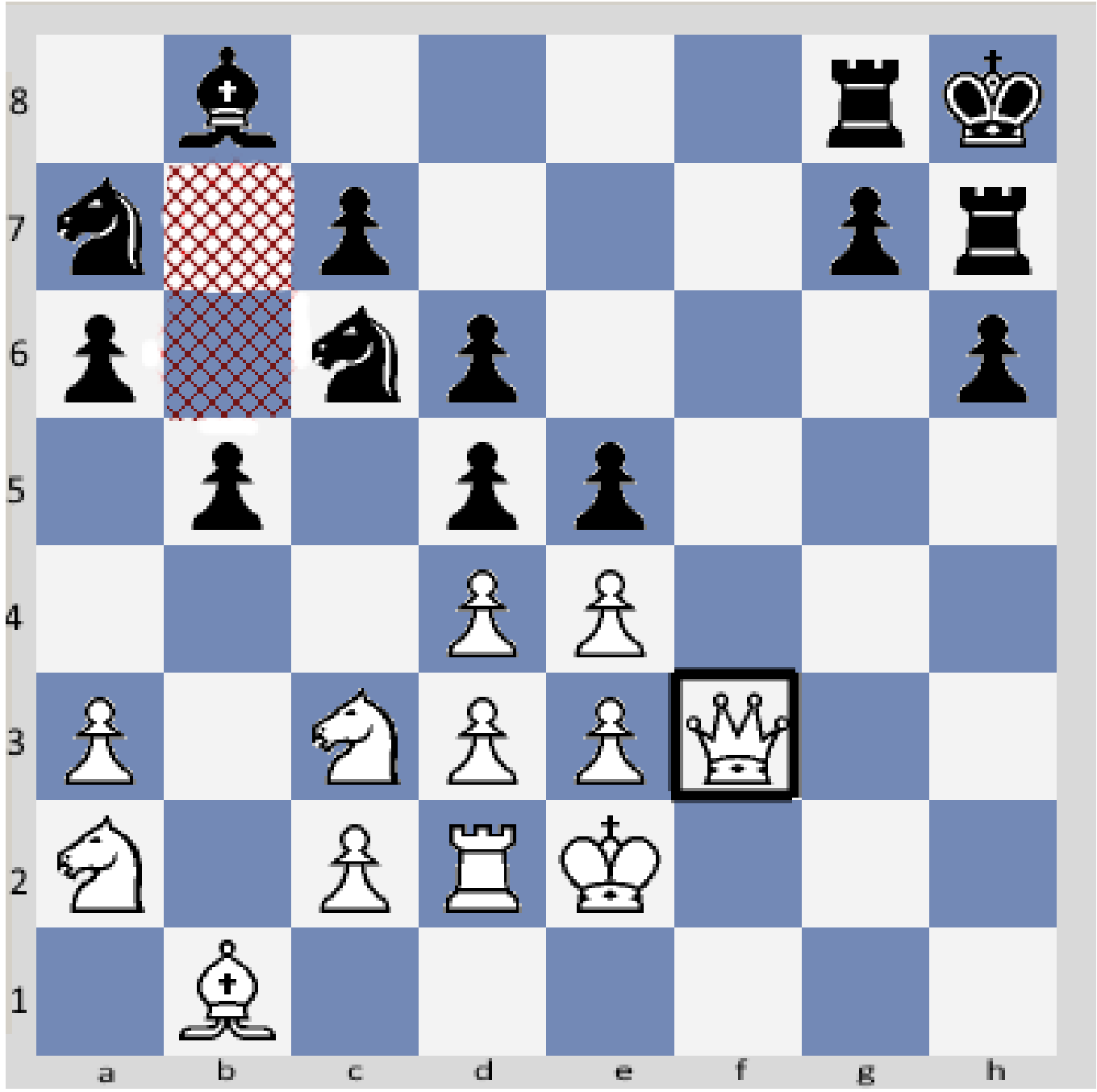

**Figure 9:** The tricky example.

Here the white queen has just moved to the f3 square (highlighted). The only possible EA on the board by SSCC rules is comprised of the b6 and b7 squares (cross hatched squares). The algorithm should first correctly detect that b6 and b7 make up the only valid EA, and then it should determine that the queen on f3 has at least two unique paths to get to that single valid EA. This test initially failed because it turned out that the first candidate path to the EA was 'cutting off' the second candidate path from reaching the EA. This was solved by a look-ahead technique. Here was the output of the algorithm on screen:

```
Board Position From FEN String:
        -       b       -       -       -       -       r       k
        n       -       p       -       -       -       p       r
        p       -       n       p       -       -       -       p
        -       p       -       p       p       -       -       -
        -       -       -       P       P       -       -       -
        P       -       N       P       P       Q       -       -
        N       -       P       R       K       -       -       -
        -       B       -       -       -       -       -       -


Classes After CCA:

        1       -       1       1       1       1       -       -
        -       1       -       1       1       1       -       -
        -       1       -       -       1       1       1       -
        1       -       1       -       -       1       1       1
        1       1       1       -       -       1       1       1
        -       1       -       -       -       -       1       1
        -       1       -       -       -       1       1       1
        1       -       1       1       1       1       1       1
```

*(the algorithm correctly finds all the empty squares 8-connected as EA group #1)*

```
Classes After CCA and Perimeter Filtering:
        -       -       -       -       -       -       -       -
        -       1       -       -       -       -       -       -
        -       1       -       -       -       -       -       -
        -       -       -       -       -       -       -       -
        -       -       -       -       -       -       -       -
        -       -       -       -       -       -       -       -
        -       -       -       -       -       -       -       -
        -       -       -       -       -       -       -       -
```

*(the perimeter filtering correctly eliminates all but b6 and b7 squares from EA group #1)*

```
Classes After CCA And Filtering:
        -       -       -       -       -       -       -       -
        -       1       -       -       -       -       -       -
        -       1       -       -       -       -       -       -
        -       -       -       -       -       -       -       -
        -       -       -       -       -       -       -       -
        -       -       -       -       -       -       -       -
        -       -       -       -       -       -       -       -
        -       -       -       -       -       -       -       -


Classes After CCA And Filtering -- Corrected:
        -       -       -       -       -       -       -       -
        -       1       -       -       -       -       -       -
        -       1       -       -       -       -       -       -
        -       -       -       -       -       -       -       -
        -       -       -       -       -       -       -       -
        -       -       -       -       -       -       -       -
        -       -       -       -       -       -       -       -
        -       -       -       -       -       -       -       -


A VALID CHAIN WAS FOUND... Try another position?   y/n :
```

*(the algorithm correctly determines there are 2 unique paths from f3 to the EA group #1)*

An even better example of this problem came up later. It was discovered that the timing of the search along each of eight candidate paths turns out to be critical. The 8-Connected Component Analysis section of the algorithm is O(N) where N is the array size (in SSCC's case, 64). This portion of the algorithm has to make 4 passes through the array to get the class components of each EA absolutely correct. However, after that comes the actual search of the eight candidate paths from the input starting square to an EA. The question here is, what is the worst case scenario? Figure 10 comes fairly close, it would seem, to the worst case scenario for the case of an 8x8 chessboard (again, please ignore the unlikelihood of this randomly-generated position).

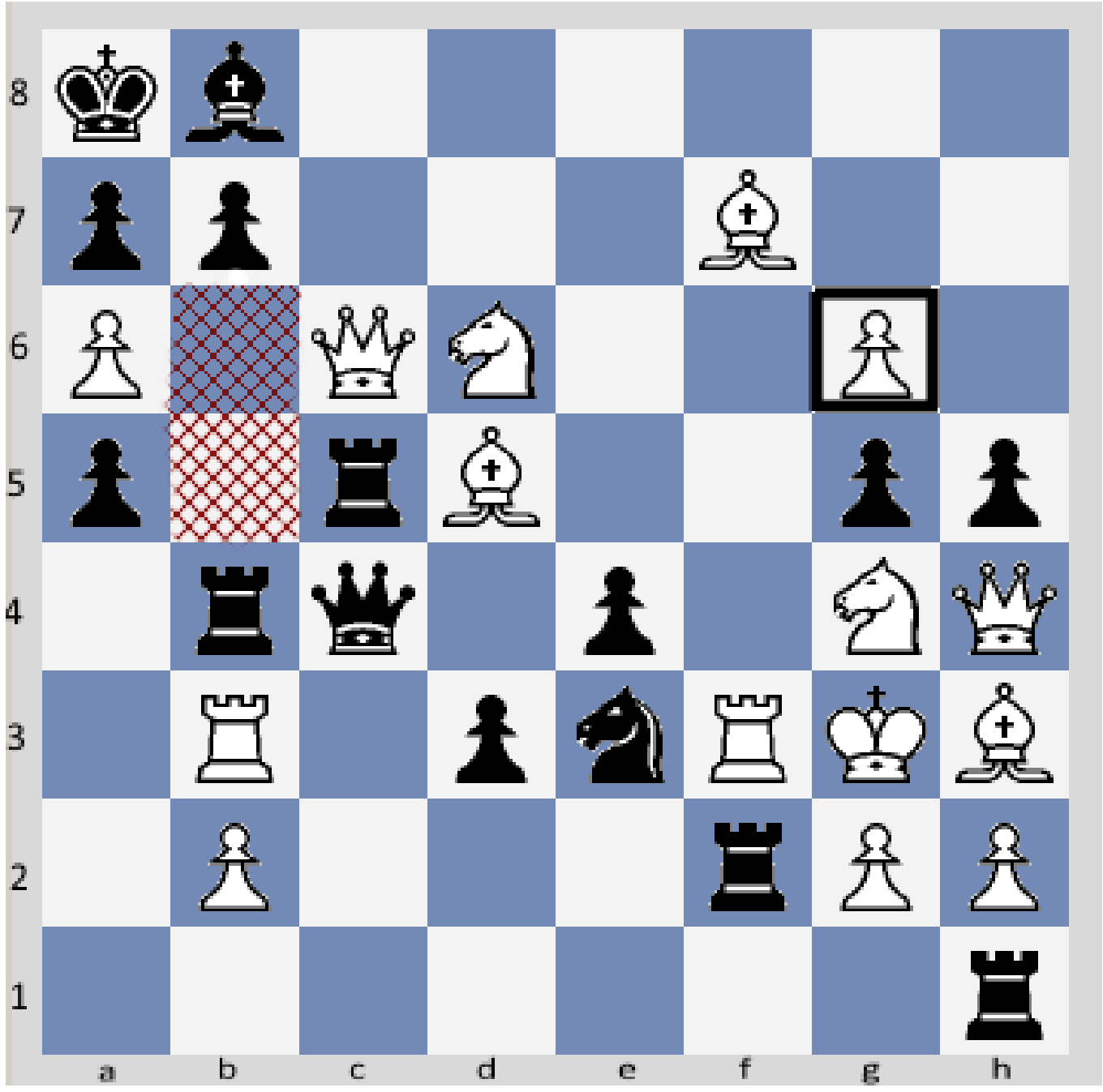

**Figure 10:** Another tricky example.

So here the sole EA on the board is comprised of the b5 and b6 squares. The last-moved piece that creates the longest search is the white pawn on the g6 square (highlighted). Note that even though the white bishop on f7 is even further away (chain-wise) from the EA, it is quickly identified as not part of a valid chain because it has only one connection, which is the pawn on g6. A human that is conditioned to identify valid chains for SSCC can quickly determine that the pawn on g6 does indeed have at least two unique paths to get to the EA. Nowhere is there a 'bottleneck' of only one piece (but if you take away any one of the white Bishop on d5 or the black Queen on c4 or the black Pawn on d3 or on e4, then yes, such a bottleneck would be in place).

The algorithm, however, is not using computer vision or pattern recognition (*which would nevertheless be an interesting path to explore in creating a competing algorithm!*) It must do a brute-force search along all candidate paths (of squares that have pieces) that can lead to either b5 or b6, and for all such paths, determine whether any two of them have no duplicate squares between them. The timing problem previously mentioned comes into play here, because once one of the eight candidate paths uses a board square, none of the others may use that square. The search order that was used to establish the eight paths from a given square (g6 in this case) is: *Top Left, Top Center, Top Right, Left, Right, Bottom Left, Bottom Center, Bottom Right*. Sometimes this search order can cause one candidate path to 'cut off' another one, causing a false negative.

It was discovered that the solution to this problem was to only mark the eight squares immediately surrounding the last-piece-moved-to square (or for non-chess applications, our specified start square) as being unable to be used by other candidate paths. Then, as each candidate path of connected squares with pieces gets square by square added to it or later removed, each added square is temporarily added to a set of visited squares and each removed square is removed from the visited squares set. This dynamic approach proved to solve the problem. It can be found in the pseudocode for the found2PathsToIndex(...) function, previously listed.

One can imagine what a near-worst-case position like the above does for execution time. The above position took many orders of magnitude longer to execute than others without such long chains of pieces (*this was only noticed in a debug build, which magnifies such differences enough to be noticed: in a release build, even this near-worst-case scenario still ran very quickly.*) The algorithm could be optimized such that on a multi-core machine or one with many GPU cores, these eight searches, each starting from a different square, could each be done on a separate physical or logical core, and the first thread to find two unique paths to a single EA could terminate all the other threads. Nevertheless, this discovery does highlight the problem of rapid growth of the time complexity with growth of the search space, which itself begs for more intelligent and optimized solutions that may be awaiting discovery.